\documentstyle[cl,lingmacros]{article}
\issue{20}{2}{1994}
\title{Japanese Discourse \\ and the Process of Centering}
\author{
\begin{tabular}{ccc}
Marilyn Walker\thanks{Computer Science Dept, 200 S. 33rd St.,Philadelphia, PA
19104, {\tt lyn@linc.cis.upenn.edu}}
& Masayo Iida\thanks{Stanford Ca. 94304, {\tt iida@csli.stanford.edu}}
& Sharon Cote\thanks{Linguistics Dept, Philadelphia, PA 19104, {\tt
cote@linc.cis.upenn.edu}} \\
University of Pennsylvania & Stanford University & University of Pennsylvania
\end{tabular}
}
\runningtitle{Japanese Discourse and the Process of Centering}
\runningauthor{Marilyn Walker and Masayo Iida and Sharon Cote}

\def\shortexdt#1#2#3#4#5{\begin{tabular}[t]{@{}*{#1}{l@{\ }}}
#2\\ #3\\ \multicolumn{#1}{@{}l@{}}{#4}\\
\multicolumn{#1}{@{}l@{}}{#5}
\end{tabular}}

\def\shortextt#1#2#3#4#5#6{\begin{tabular}[t]{@{}*{#1}{l@{\ }}}
#2\\ #3\\ \multicolumn{#1}{@{}l@{}}{#4}\\
\multicolumn{#1}{@{}l@{}}{#5}\\
\multicolumn{#1}{@{}l@{}}{#6}
\end{tabular}}

\begin{document}           
\maketitle                 
\bibliographystyle{fullname}

\begin{abstract}

This paper has three aims: (1) to generalize a computational account
of the discourse process called {\sc centering}, (2) to apply this
account to discourse processing in Japanese so that it can be used in
computational systems for machine translation or language
understanding, and (3) to provide some insights on the effect of
syntactic factors in Japanese on discourse interpretation.  We argue
that while discourse interpretation is an inferential process,
syntactic cues constrain this process, and demonstrate this argument
with respect to the interpretation of {\sc zeros}, unexpressed
arguments of the verb, in Japanese.  The syntactic cues in Japanese
discourse that we investigate are the morphological markers for
grammatical {\sc topic}, the postposition {\it wa}, as well as those
for grammatical functions such as {\sc subject}, {\em ga}, {\sc
object}, {\em o} and {\sc object2}, {\em ni}.  In addition, we
investigate the role of speaker's {\sc empathy}, which is the
viewpoint from which an event is described.  This is syntactically
indicated through the use of verbal compounding, i.e.  the auxiliary
use of verbs such as {\it kureta, kita}.  Our results are based on a
survey of native speakers of their interpretation of short discourses,
consisting of minimal pairs, varied by one of the above factors. We
demonstrate that these syntactic cues do indeed affect the
interpretation of {\sc zeros}, but that having previously been the
{\sc topic} and being realized as a {\sc zero} also contributes to the
salience of a discourse entity.  We propose a discourse rule of {\sc
zero topic assignment}, and show that {\sc centering} provides
constraints on when a {\sc zero} can be interpreted as the {\sc zero
topic}.
\end{abstract}

\section{Introduction}
\subsection{Centering in Japanese Discourse}
\label{intro-sec}

Recently there has been an increasing amount of work in computational
linguistics involving the interpretation of anaphoric elements in
Japanese  \cite{Yoshimoto88,Kuno89,WIC90,Nakagawa92}.  These accounts
are intended as components of computational systems for machine
translation between Japanese and English or for natural language
processing in Japanese alone. This paper has three aims: (1) to
generalize a computational account of the discourse process called
{\sc centering} (Sidner, 1979; Joshi and Weinstein, 1981; Grosz, Joshi
and Weinstein, 1983; Grosz, Joshi and Weinstein, 1986), (2) to apply
this account to discourse processing in Japanese so that it can be
used in computational systems, and (3) to provide some insights on the
effect of syntactic factors in Japanese on discourse interpretation.

In the computational literature, there are two foci for research on
the interpretation of anaphoric elements such as pronouns.  The first
viewpoint focuses on an inferential process driven by the underlying
semantics and relations in the domain  \cite{Hobbs86,HCDEL87,HM87}.  A
polar focus is to concentrate on the role of syntactic information
such as what was previously the topic or subject
 \cite{Hobbs76a,Kameyama85,Yoshimoto88}.  We will argue for an
intermediate position with respect to the interpretation of {\sc
zeros}, unexpressed arguments of the verb, in Japanese.  Our position
is that the interpretation of zeros is an inferential process, but
that syntactic information provides constraints on this inferential
process  \cite{JK79,JW81}. We will argue that syntactic
cues and semantic interpretation are mutually constraining
 \cite{Prince81,Prince85,Hudson88}.

The syntactic cues in Japanese discourse that we investigate
are the morphological markers for grammatical {\sc topic},
the postposition {\it wa}, as well as those for grammatical
functions such as {\sc subject}, {\em ga}, {\sc object},
{\em o} and {\sc object2}, {\em ni}.  In addition, we
investigate the role of speaker's {\sc empathy}, which is
the viewpoint from which an event is described. This can be
syntactically indicated through the use of verbal
compounding, i.e.  the auxiliary use of verbs such as {\it
kureta, kita}.

In addition to the argument that a purely inference-based account does
not consider limits on processing time, another argument against a
purely inference-based account is provided by the minimal pair below.
Here, the only difference is whether Ziroo is the subject or the
object in the second utterance.  Note that the interpretation of zeros
is indicated in parentheses:

\eenumsentence{\item[a.]
\shortex{5}
{Taroo &ga &kooen &o &sanpositeimasita.}
{Taroo &{\sc subj}&park&in&walking-was}
{\it Taroo was taking a walk in the park.}

\item[b.]
\shortex{8}
{Ziroo &ga &0 &hunsui&no &mae &de &mitukemasita.}
{Ziroo&{\sc subj}&{\sc obj}&fountain&of&front&in&found}
{\it Ziroo found (Taroo)in front of the fountain.}

\item[c.]
\shortex{10}
{0&0&kinoo&no &siai &no &kekka &o &kikimasita.}
{{\sc subj}&{\sc obj}&yesterday &of &game &of &scores&{\sc obj}&asked}
{\it (Ziroo) asked (Taroo) the score of yesterday's game.}
}

\label{ex-1}

\eenumsentence{\item[a.]
\shortex{5}
{Taroo &ga &kooen &o &sanpositeimasita.}
{Taroo &{\sc subj}&park&in&walking-was}
{\it Taroo was taking a walk in the park.}

\item[b.]
\shortex{8}
{0&Ziroo&o &hunsui&no &mae &de &mitukemasita.}
{{\sc subj}&Ziroo&{\sc obj}&fountain&of&front&in&found}
{\it (Taroo) found Ziroo in front of the fountain.}

\item[c.]
\shortex{10}
{0&0&kinoo&no &siai &no &kekka &o &kikimasita.}
{{\sc subj}&{\sc obj}&yesterday &of &game &of &scores&obj&asked}
{\it (Taroo) asked (Ziroo) the score of yesterday's game.}
}
\label{ex-2}

In \ex{-1}b and \ex{0}b, the syntactic position in which Ziroo
is realized has the effect that \ex{-1}c means {\it Ziroo asked Taroo
the score of yesterday's game}, while \ex{0}c means {\it Taroo asked
Ziroo the score of yesterday's game}. On the other hand, some purely
syntactic accounts require that antecedents for zeros be realized as
the grammatical {\sc topic}, and thus cannot explain the above example because
Taroo is never explicitly marked as the topic
 \cite{Yoshimoto88}.

In the literature, {\sc zeros} are known as zero pronouns. We adopt
the assumption of earlier work that the interpretation of zeros in
Japanese is analogous to the interpretation of overt pronouns in other
languages  \cite{Kuroda65,Martin76,Kameyama85}.  Japanese also has
overt pronouns, but the use of the overt pronoun is rare in normal
speech, and is limited even in written text.  This is mainly because
overt pronouns like {\it kare\/} (`he') and {\it kanozyo\/} (`she')
were introduced into Japanese in order to translate gender-insistent
pronouns in foreign languages  \cite{Martin76}.  In this paper, we only
consider zeros in subcategorized-for argument positions.  Since
Japanese doesn't have subject or object verb agreement, there is no
syntactic indication that a zero is present in an utterance other
than information from subcategorization.\footnote{When zero pronouns
should be stipulated is still a research issue. For example,
 \cite{Hasegawa84} described a zero pronoun as a phonetically null
element in an argument position.  However, as shown in the following
example,
 \cite{TYI80} assumed that zero pronouns are not
limited in their distribution and stipulated them in
adjunct positions as well \cite{Iida92}.

\shortex{7}
{Taroo & wa& Hanako & no &  kaban & o & mitukemasita.}
{Taroo &{\sc top/subj}&Hanako&{\sc gen}&bag&{\sc obj}&found}
{\it Taroo found Hanako's bag.}

\shortex{7}
{0 & 0& tanzyoobi& no & purezento & o & iremasita.}
{ & & birthday & {\sc gen} &present & {\sc obj} & put}
{{\it (Taroo) put a birthday present (in her bag).}}
}

First, in section \ref{method-sec} we describe the
methodology that we applied in this investigation.  In
section \ref{cent-sec}, we present the theory of Centering
and some illustrative examples.  Then, in section
\ref{jap-cent-sec}, we discuss particular aspects of
Japanese discourse context, namely grammatical {\sc topic} and
speaker's {\sc empathy}. We will show how these can easily be
incorporated into a centering account of Japanese discourse
processing, and give a number of examples to illustrate the
predictions of the theory.  We also discuss the way in which a
discourse center is instantiated in section \ref{cb-instant-sec}.

In section \ref{zta-sec} we propose a discourse rule of {\sc
zero topic assignment}, and use the centering model to formalize
constraints on when a zero may be interpreted as a {\sc zero topic}.
Our account makes a distinction between two notions of {\sc topic},
grammatical topic and zero topic.  The grammatical topic is the {\it
wa}-marked entity, which is by default predicted to be the most
salient discourse entity in the following discourse.  However there
are cases in which it may not be, depending on whether {\sc zero topic
assignment} applies. This analysis provides support for Shibatani's
claim that the interpretation of the topic marker, {\it wa}, depends
on the discourse context  \cite{Shibatani90}.  {\sc zero topic
assignment} actually predicts ambiguities in Japanese discourse
interpretation and provides a mechanism for deriving interpretations
that previous accounts claim would be unavailable.

We delay the review of related research to section \ref{rel-res-sec}
when we can contrast it with our account.  The two major previous
accounts are those of Kuno  \cite{Kuno72,Kuno76a,Kuno87,Kuno89} and
Kameyama  \cite{Kameyama85,Kameyama86b,Kameyama86a}. Finally in section
\ref{conclusion-sec}, we summarize our results and suggest topics for
future research.

\subsection{Methodology}
\label{method-sec}

Most of the examples in this paper are constructed as four
utterance discourses that fit one of a number of structural
paradigms.  In all of the paradigms, a discourse entity is
introduced in the first utterance, and established by the
second utterance as the {\sc center}, what the discourse is
about.  The manipulations of context occur with
the third and the fourth utterances. In each case the zero
in the third utterance cospecifies the entity already
established as the center in the second
utterance.  The fourth utterance consists of a potentially
ambiguous sentence containing two zeros.
The variations in context are as shown below:
\begin{center}
\begin{tabular}{|lc|cc|c|}
\hline  &&&&\\
\multicolumn{2}{c}{Third Utterance} & \multicolumn{2}{c}{Fourth Utterance} &
\multicolumn{1}{c}{}\\
{\sc subject} & {\sc object(2)} & {\sc subject} & {\sc object(2)} &
{\sc examples} \\
\hline  &&&&\\
 zero & NP(o or ni) & zero & zero & \ref{cont-ret-ex} \\ \hline \\
 zero & NP(o or ni) & zero & zero, empathy & \ref{zta-emp-cont} \\
\hline  &&&&\\
 NP(ga) & zero & zero & zero & \ref{zta-ex-ga}, \ref{zta-emp-ga-noemp} \\
\hline  &&&&\\
 NP(wa) & zero & zero & zero & \ref{shift-ex}, \ref{zta-ex-wa}  \\
\hline  &&&&\\
 NP(ga) & zero & zero & zero, empathy & \ref{zta-emp-ga} \\ \hline
\end{tabular}
\end{center}

Thus we are manipulating factors such as whether a
discourse entity is realized in subject or object
position in the third utterance, whether a discourse
entity realized in subject position is {\it ga}-marked
or {\it wa}-marked in the third utterance, and whether
a discourse entity realized in the fourth utterance in
object position is marked as the locus of speaker's
{\sc empathy}.

We collected a group of about 35 native speakers by
solicitation on the net to provide judgements for most of
the examples given in this paper.  These native speakers
were readers of the newsgroups sci.lang.japanese and
comp.research.japan. They were thus typically well-educated,
bilingual engineers.  Whenever an example was tested in
this way, we will provide the number of informants who
chose each possible interpretation to the right of the
example. Some examples that are included for expository
reasons were not tested.

Participation in our survey was completely voluntary
and the data was collected over 3 surveys. Thus the
numbers of subjects varied from one survey to another
and this is reflected in the numbers accompanying our
examples.  This data collection was carried out on
written examples using electronic mail in a situation
in which the informants could take as long as they
wanted to decide which interpretation they preferred.
The instructions sent with the surveys are given in the
appendix.

This paradigm clearly cannot provide information on which
interpretation a subject might arrive at first and then
perhaps change based on other pragmatic factors, and thus it
contrasts with reaction time studies. However the judgements
given
should be stable, and reflect the fact that our
informants were able to use all the information in the
discourse. It is a useful paradigm given that we are
exploring the correlation of syntactic cues and discourse
interpretation. It has been claimed that syntactic cues are
only used in automatic processing and can be over-ridden by
deeper processing. However Hudson's results suggest that
subjects may judge a discourse sequence to be nonsensical
when it is incoherent according to centering  \cite{Hudson88},
Chap5. Di Eugenio claims that discourse sequences in Italian
that are not discourse coherent according to centering
theory produce a garden-path effect  \cite{Dieugenio90}. The
methods we used allow us to explore the results of these
interactions, and yet it would be beneficial for these
results to be expanded upon by careful psychological
experimentation \cite{HudsonTanenhaus95}.

For most of the examples reported here, we asked subjects to
choose one preferred interpretation instead of allowing
them to rank interpretations. The motivation for doing this
was to force differences to come out for slight preferences,
with the theory being that other variations would come out
across subjects. In a few cases we allowed subjects to
indicate no preference; these examples will be clearly
indicated.

In addition, we used the same gender for multiple discourse
entities to prevent any tendency for judgements to be
influenced by gender stereotypes. We also avoided using
verbs with causal biases toward one of their arguments, and
we used few cue words such as {\it but, because, then},
which could result in a bias towards, say, a cause-effect or
temporal sequence of events interpretation.  We also omitted
honorific markers, which are normally a part of Japanese
ambiguity resolution.\footnote{While native speakers
understandably found some of these examples ``stilted'' or
``awkward'', they were still able to give their judgements
based on the information that was provided in the
discourses.} This was done to isolate the effects of the
variables that we were exploring in this study, namely topic
marking, grammatical function, empathy, and realization with
a zero or with a full noun phrase.

\section{Centering Theory}
\label{cent-sec}

Within a theory of discourse, {\sc centering} is a
computational model of the process by which conversants
coordinate attention in discourse  \cite{GJW86}.  Centering has
its computational foundations in the work of Grosz and Sidner
 \cite{Grosz77,Sidner79,GS85} and was further developed by
Grosz, Joshi and Weinstein  \cite{GJW83,GJW86,JW81}.  Centering
is intended to reflect aspects of {\sc attentional state} in a
tripartite view of discourse structure that also includes {\sc
intentional structure} and {\sc linguistic structure}
 \cite{GS86}.  In Grosz and Sidner's theory of discourse
structure, discourses can be segmented based on intentional
structure and a discourse segment exhibits both local and
global coherence.  Global coherence depends on how each
segment relates to the overall purpose of the discourse; local
coherence depends on aspects such as the syntactic structure
of the utterances in that segment, the choice of referring
expressions, and the use of ellipses.  {\sc centering} models
local coherence and is formalized as a system of constraints
and rules.  Our analysis uses an adaptation of a centering
algorithm that was developed by Brennan, Friedman and Pollard,
based on these constraints and rules (Brennan, Friedman and
Pollard,1987),  \cite{Walker89b}.

The purpose of centering as part of a computational model of discourse
interpretation is to model {\sc attentional state} in discourse in
order to control inference  \cite{JK79,JW81}.\footnote{Recent work in
situation theory proposes to control computation with a similar notion
of background information in terms of constants of the situation that
thus are not explicitly realized in an utterance  \cite{Nakashima90}.
The situation-theoretic work does not as yet distinguish shared
knowledge that determines discourse salience and derives from the
discourse context and the way utterances are expressed
 \cite{CH77,CM81,Prince81} from shared knowledge that is part of
general background knowledge such as cultural assumptions
 \cite{Prince78b,Joshi82} or shared knowledge that might derive from
the task context  \cite{Grosz77}.} Our approach to modeling attentional
state is to explore aspects of the correlation between syntax and
discourse function.  This assumes that there are language conventions
about discourse salience and that conversants attempt to maintain a
sense of shared context.

Section \ref{rules-sec} presents the centering rules and
constraints. Section \ref{cont-ret-sec} and \ref{shift-sec}
illustrate the theory and the definitions with a number of
examples. Section \ref{cent-alg-sec} discusses the centering
algorithm for the resolution of zeros in Japanese.

\subsection{Rules and Constraints}
\label{rules-sec}

The centering model is very simple.  Each utterance in a discourse
segment has two structures associated with it.  First, each utterance
in a discourse has associated with it a set of discourse entities
called {\sc forward-looking centers}, $\rm{Cf}$. Centers are semantic
entities that are part of the discourse model.  Second, there is a
special member of this set called the {\sc backward-looking center},
$\rm{Cb}$. The $\rm{Cb}$ is the discourse entity that the utterance
most centrally concerns, what has been elsewhere called the `theme'
 \cite{Reinhart81,Horn86}. The $\rm{Cb}$ entity links the current
utterance to the previous discourse.

The set of {\sc forward-looking centers}, Cf, is ranked according to
discourse salience. We will discuss factors that determine the ranking
below. The highest ranked member of the set of forward looking centers
is referred to as the {\sc preferred center}, $\rm{Cp}$.\footnote{The
notion of {\sc preferred center} corresponds to Sidner's notion of
{\sc expected focus}  \cite{Sidner83a}.} The {\sc preferred center}
represents a prediction about the Cb of the following utterance.
Sometimes the Cp will be what the previous segment of discourse was
about, the Cb, but this is not necessarily the case. This distinction
between looking back to the previous discourse with the Cb and
projecting preferences for interpretation in subsequent discourse with
the Cp is a key aspect of centering theory.

In addition to the structures for centers, Cb and Cf, the theory
of centering specifies a
set of rules and constraints. Constraints are meant to hold
strictly whereas rules may sometimes be violated.
\begin{itemize}
\item
{\bf CONSTRAINTS} \\
For each utterance $\rm{U_{i}}$ in a discourse segment $\rm{U_{1}, \ldots
,U_{m}}$:
   \begin{enumerate}
      \item There is precisely one backward looking center $\rm{Cb}$.
      \item Every element of the forward centers list,
$\rm{Cf(U_{i}})$, must be realized in $\rm{U_{i}}$.
      \item The center, $\rm{Cb(U_{i})}$, is the highest-ranked element
of $\rm{Cf(U_{i-1})}$ that
                 is realized in $\rm{U_{i}}$.\footnote{This could
possibly be rephrased as: Assume the $\rm{Cp(U_{i-1}}$
is the $\rm{Cb(U_{i})}$ unless there is evidence to the
contrary \cite{Carter87}.}
    \end{enumerate}
\end{itemize}

Constraint (1) says that there is one central discourse entity that
the utterance is about, and that is the $\rm{Cb}$. The second
constraint depends on the definition of {\em realizes\/}.  An
utterance U {\em realizes\/} a center {\bf c} if {\bf c} is an element
of the situation described by U, or {\bf c} is the semantic
interpretation of some subpart of U \cite{GJW86}.  Thus the relation
{\sc realize} describes zeros, explicitly realized discourse entities,
and those implicitly realized centers that are entities inferable from
the discourse situation  \cite{Prince78b,Prince81}.

A specialization of the relation {\sc realize} is the
relation {\sc directly realize}. A center is directly
realized if it corresponds to a phrase in an utterance. We
restrict our focus to entities realized by noun phrases,
however it is clear that propositions can be centers, so we
assume that the account given here can be extended to
propositional entities as
well  \cite{Webber78,Sidner79,Prince78a,Ward85,Prince86}.

As we discuss further in section \ref{jap-cent-sec}, zeros refer to
entities that are already in the discourse context.  The fact that the
current utterance {\sc realizes} one or more zeros follows from
information specified in the subcategorization frame of the verb.
These arguments must be interpreted and thus acquire a degree of
discourse salience that nonsubcategorized-for discourse entities lack.

Constraint (3) stipulates that the ranking of the forward centers,
$\rm{Cf}$, determines from among the elements that are realized in the
next utterance, which of them will be the $\rm{Cb}$ for that
utterance. If the {\sc preferred center}, $\rm{Cp(U_i)}$, is realized
in $\rm{U_{i+1}}$, it is predicted to be the $\rm{Cb(U_{i+1})}$.  We
will use the following forward center ranking for
Japanese:\footnote{This ranking is consistent with Kuno's Empathy
Hierarchies and with Kameyama's Expected Center Order
 \cite{Kuno87,Kameyama85,Kameyama86a}.  This will be discussed in
section \ref{rel-res-sec}.  We do not include discourse entities for
verb phrases or other propositional entities in this ranking since we
have not studied their contribution, but see
 \cite{Sidner79,Sidner81,Carter87}.}

\begin{quote}
{\sc (grammatical OR zero) topic} $>$ {\sc empathy} $>$ {\sc subject}
$>$ {\sc object2} $>$ {\sc object} $>$ {\sc others}
\end{quote}

Backward-looking centers, Cbs, are often deleted or pronominalized and
some transitions between discourse segments are more coherent than
others.  According to the theory of centering, coherence is measured
by the hearer's inference load when interpreting a discourse
sequence \cite{JW81,GJW86}.  For instance, discourse segments that continue
centering the same entity are more coherent than those that repeatedly
shift from one center to another.  These observations are encapsulated
in two rules:

\begin{itemize}
\item
{\bf RULES} \\
For each $\rm{U_{i}}$ in a discourse segment $\rm{U_{1}, \ldots ,U_{m}}$:
   \begin{enumerate}
     \item If some element of $\rm{Cf(U_{i-1})}$ is realized as a
pronoun in $\rm{U_{i}}$,
              then so is $\rm{Cb(U_{i})}$.
     \item Transition states are ordered.
   {\sc continue\/} is preferred to {\sc retain\/} is preferred to
 {\sc smooth-shift} is preferred to {\sc rough-shift}.\footnote{
Smooth-shift was called shifting-1 by  \cite{BFP87}.}
\end{enumerate}
\end{itemize}

Rule (1) captures the intuition that pronominalization is
one way to indicate discourse salience. It follows from
Rule (1) that if there are multiple pronouns in an
utterance, one of these must be the Cb.  In addition, if
there is only one pronoun, then that pronoun must be the
Cb.  For Japanese, we extend this rule directly to zeros,
assuming that
zeros in Japanese correspond to destressed pronouns in
English.

Rule (2) states that modeling attentional state depends on analyzing
adjacent utterances according to a set of transitions that measure the
coherence of the discourse segment in which the utterance occurs.
Measuring coherence is based on an estimate of the hearer's inference
load, but this measure must always relative since there is no grammar
of discourse. Thus methods for exploring these issues must
use comparative measures of how some
discourses are easier to process than others. Centering Theory
models this by stipulating that some transitions are preferred over
others.

The typology of transitions from one utterance, $\rm{U_{i}}$, to the
next, is based on two factors: whether the backward-looking center,
$\rm{Cb}$, is the same from $\rm{U_{i-1}}$ to $\rm{U_{i}}$, and
whether this discourse entity is the same as the preferred center,
$\rm{Cp}$, of $\rm{U_{i}}$.\footnote{It is possible that restricting
the relation between the $\rm{Cb(U_i)}$ and the $\rm{Cb(U_{i-1})}$ to
be coreference (equality) may be too strong. Future work should
examine the role of shifts to functionally dependent entities or
entities related by poset relations to the previous Cb.}

\begin{enumerate}
\item $\rm{Cb(U_{i}) = Cb(U_{i-1})}$, or there is no $\rm{Cb(U_{i-1})}$
\item $\rm{Cb(U_{i}) = Cp(U_{i})}$
\end{enumerate}

If both (1) and (2) hold then we are in a {\sc continue} transition.
The {\sc continue} transition corresponds to cases where the speaker
has been talking about a particular entity and indicates an intention
to continue talking about that entity.\footnote{A prediction made by
the preference for {\sc continue} is that intersentential antecedents
for pronouns will be preferred over intrasentential candidates.  This
preference is one that distinguishes Centering for pronoun
interpretation from the proposal made by Hobbs in
 \cite{Hobbs76a,Hobbs76b}.  However this preference needs to be
constrained further by the fact that sortal filters may rule out the
Cp of the previous utterance as the current Cb.  In this case the data
suggests that perhaps intrasentential candidates should be preferred
 \cite{Walker89b}. Carter explored this in his extension of Sidner's
theory of local focusing  \cite{Carter87}.} If (1) holds but (2)
doesn't hold then we are in a {\sc retain} transition. {\sc retain}
corresponds to a situation where the speaker is intending to {\sc
shift} onto a new entity in the next utterance and is signalling this
by realizing the current center in a lower ranked position on the Cf
(examples follow below).

If (1) doesn't hold then we are in one of the {\sc shift}
states depending on whether or not (2) holds. This
definition of transition states is summarized in Figure
\ref{state-fig}  \cite{BFP87}.  We will use the notation of
$\rm{Cb(U_{i-1}) =}$ [?] for cases where there is no
$\rm{Cb(U_{i-1})}$. Section \ref{cb-instant-sec} will discuss
center instantiation.

\begin{figure}[htb]
\begin{center}
\begin{tabular}{r|c|c}
& $\rm{Cb(U_i) = Cb(U_{i-1})}$ & $\rm{Cb(U_i) \neq Cb(U_{i-1})}$
\\
& OR $\rm{Cb(U_{i-1}) =}$ [?] \\ \hline
& & \\
$\rm{Cb(U_i) = Cp(U_i)}$ & {\sc continue} & {\sc smooth-shift} \\ \hline
& & \\
$\rm{Cb(U_i) \neq Cp(U_i)}$ & {\sc retain} & {\sc rough-shift} \\ \hline
\end{tabular}
\caption{Centering Transition States, Rule 2}
 KEY \\
{\sc backward-looking center} $= $ Cb \\
{\sc preferred center} $=$ Cp \\
Uninstantiated Cb $=$ {[?]}
\end{center}

\label{state-fig}
\end{figure}

The combination of the constraints, rules and transition states makes
a set of testable predictions about which interpretations hearers will
prefer because they require less processing. For example, maximally
coherent segments are those that require less processing time. A
sequence of a {\sc continue} followed by another {\sc continue} should
only require the hearer to keep track of one main discourse entity,
which is currently both the Cb and the Cp. A single pronoun in an
utterance is the current Cb (by Rule 1) and can be interpreted to
cospecify the discourse entity realized by $\rm{Cp(U_{i-1})}$ in one
step(Constraint 3).

The ordering of the Cf is the main determinant of which transition
state holds between adjacent utterances.  This means that the
predictions of the theory are largely determined by the ranking of the
items on the $\rm{Cf}$.  But there are many factors that can
contribute to the salience of a discourse entity; among them are
factors that we will not examine here such as lexical semantics,
intonation, word-order, and tense.\footnote{See  \cite{Hudson88} for an
examination of the role of lexical semantics in Centering.} In this
paper we explore the influence of various syntactic factors, which we
discuss in detail in section \ref{jap-cent-sec}.  We will also examine
the relative contribution of pronominalization and postposition
marking in section \ref{zta-sec}.  We postulate that the Cf
ordering will vary from language to language depending on the means
the language provides for expressing discourse function.  However much
of this variation can be captured in the ranking of the $\rm{Cf}$ due
to the modularity of the theory.

In sections \ref{cont-ret-sec} and \ref{shift-sec} we will present
some simple examples to motivate these definitions.  In section
\ref{cent-alg-sec} we will present a slightly modified version of the
centering algorithm \cite{BFP87}. In the following discussion we
assume that the centering rules and constraints, and the notion of
centering transition states have some cognitive reality
\cite{Brennan93,Hudson88,GGG93,HudsonTanenhaus95}. However we make no
claims about the cognitive reality of the centering algorithm that we
discuss in section \ref{cent-alg-sec}.

\subsection{The Distinction between Continue and Retain}
\label{cont-ret-sec}

This theory predicts preferences in the interpretation of
utterances whose meaning depends on parameters from the
discourse context.  Thus if there are still multiple
possibilities for interpretation after the application of
all constraints and rules, the ordering on transitions
applies, and {\sc continue} interpretations are
preferred(Rule 2).  Indeed, many cases of the preference
for one interpretation over another follow directly from
the distinction between the transition states of {\sc
continue} and {\sc retain}.  Let us look at a simple
example. In the discourse segment in
\ex{1}: the zero in the second sentence is
understood as referring to {\it Taroo\/}, and not to {\it
Hanako\/}. Remember that the interpretation of zeros is
indicated with parentheses.

\eenumsentence{\item[a.]
\shortex{7}{Taroo&wa&Hanako&o&eiga&ni&sasoimasita.}
         {Taroo&{\sc top/subj}&Hanako&{\sc obj}&movie&to&invited}
          {{\it Taroo invited Hanako to the movie.}}

\begin{tabular}{|llllll|}
\hline
{\bf Cb:} & {\sc taroo} & & & & \\
{\bf Cf:} & [{\sc taroo}, {\sc hanako}] & & & & \\ \hline
\end{tabular}

\item[b.]
\shortex{8}
{0 &itiniti-zyuu &nani&mo &te &ni&tukimasendesita.}
{{\sc subj}  &all-day &anything&even &hand &to &attached-not}
{{\it (Taroo)  could not do anything all day.}}

\begin{tabular}{|llllll|}
\hline
{\bf Cb:} & {\sc taroo} & & & & \\
{\bf Cf:} & [{\sc taroo}] & & & & \\ \hline
\end{tabular}
}

In example \ex{0}, the Cf from \ex{0}a contains the discourse
entity for Taroo as the first element and for Hanako as the second
element. When the unexpressed argument is interpreted in \ex{0}b, the
information from this Cf is used.  Because the zero subject may
{\sc realize} either Taroo or Hanako, both Constraint 3 and Rule 1
would be obeyed with either interpretation.\footnote{The hypothesis
that {\it wa} in \ex{0}a instantiates Taroo as the Cb will be
discussed in section \ref{cb-instant-sec}.} However by interpreting
the zero as Taroo, Taroo is the Cb, and it is possible to get a
preferred {\sc continue} interpretation {\it Taroo could not do
anything all day\/}. In this interpretation, Taroo is both the
Cb(\ex{0}b) and the Cp(\ex{0}b).

\subsection{The Distinction between Smooth-Shift and Rough-Shift}
\label{shift-sec}

In example \ex{1}, we illustrate the difference between the transition
states of {\sc rough-shift} and {\sc smooth-shift}.  Remember that
{\sc rough-shift} is claimed to be less coherent than {\sc
smooth-shift}  \cite{BFP87}.  In both cases the speaker has shifted
the center to a different discourse entity.  However in the {\sc
smooth-shift} transition state, the speaker has indicated an intention
to continue talking about the recently shifted-to entity by realizing
that entity in a highly ranked Cf position such as subject,
whereas no such indication is available with the {\sc rough-shift}
transition.  The numbers shown to the right of an interpretation
correspond to how many native speakers preferred that interpretation.


\eenumsentence{\item[a.]
\shortex{7}
{Taroo &ga &kooen &de &hon &o &yondeimasita.}
{Taroo&{\sc subj} &park &at &book &{\sc obj}&reading-was }
{{\it Taroo was reading a book in the park.}}

\begin{tabular}{|llllll|}
\hline
{\bf Cb:} & [?]   & & & & \\
{\bf Cf1:} & [{\sc taroo}, & {\sc book}] &  & & \\
& {\sc subj} & {\sc obj} & & & \\  \hline
\end{tabular}

\item[b.]
\shortex{8}
{0 &koora &o &kai &ni &baiten &ni &hairimasita.}
{{\sc subj} &cola &{\sc obj} &buy &to &shop &into &entered}
{\it { (Taroo) entered a shop to buy a cola. }}

\begin{tabular}{|lllll|}
\hline
{\bf Cb:} & {\sc taroo}   & & &  \\
{\bf Cf1:} & [{\sc taroo}, & {\sc cola}] & {\sc continue} &  \\
& {\sc subj} & {\sc obj} & &  \\  \hline
\end{tabular}

\item[c.]
\shortex{6}
{Ziroo &wa &0 &sokode &guuzen &dekuwasimasita.}
{Ziroo  &{\sc top/subj} & {\sc obj} &there &by chance &met }
{\it {Ziroo met (Taroo) there by chance. }}

\begin{tabular}{|lllll|}
\hline
{\bf Cb:} & {\sc taroo}   &  & & \\
{\bf Cf:} & [{\sc ziroo},& {\sc taroo}] & {\sc retain} &  \\
& {\sc top} & {\sc obj} & &  \\  \hline
\end{tabular}

\item[d.]
\shortex{5}
{0 &0 &eiga &ni &sasoimasita.}
{{\sc subj}&{\sc obj}  &movie  &to &invited.}
{{\it (Ziroo) invited (Taroo) to a movie.}}

\begin{tabular}{|llllll|}
\hline
{\bf Cb:} & {\sc ziroo}   & & & & \\
{\bf Cf1:} & [{\sc ziroo},& {\sc taroo}] & {\sc smooth-shift} & & 32\\
& subj & obj & & & \\  \hline
{\bf Cf2:} & [{\sc taroo},& {\sc ziroo }] & {\sc rough-shift} & & 2\\
& {\sc subj} & {\sc obj} & & & \\ \hline
\end{tabular}
\label{shift-ex}
}

In example \ex{0}, the use of {\sc topic} marking in the phrase {\it
Ziroo wa} of utterance (c) means that (c) is interpreted as a {\sc
retain}.\footnote{It has also been claimed that symmetric verbs such
as {\it meet by chance} mark {\sc empathy} on the
subject \cite{Kuno76b}.} Ziroo becomes the most highly ranked
discourse entity for c, although Taroo is the Cb since
Taroo was most highly ranked for utterance (b) (by Constraint 3).
Then when we apply the Centering algorithm in (d), there are two
candidates for the Cb(d) from the Cf(c), both Ziroo and
Taroo.  However this time when constraint 3 applies, stipulating that
the Cb must be the highest ranked element of Cf(c) realized in
\ex{0}d, Ziroo must be the highest ranked entity realized, and
therefore must be the Cb. At this point it is clear that some kind of
{\sc shift} is forced by the application of constraint 3. The two
candidates are a {\sc smooth-shift} and a {\sc rough-shift}.  The {\sc
smooth-shift} interpretation corresponds to the reading {\it Ziroo
invited Taroo to a movie} whereas the {\sc rough-shift} interpretation
corresponds to the {\it Taroo invited Ziroo} reading. The {\sc
smooth-shift} interpretation is more highly ranked, thus considered
more coherent and so is the preferred interpretation(Z =10.93, p $<$
.001).

\subsection{The Centering Algorithm}
\label{cent-alg-sec}

The {\sc centering algorithm} that was proposed by Brennan, Friedman
and Pollard incorporates the centering rules and constraints in
addition to contra-indexing constraints on coreference
 \cite{Reinhart76,BFP87,Iida92}.  These contra-indexing constraints
specify that in a sentence such as {\it He likes him}, that {\it he}
and {\it him} cannot co-specify the same discourse entity.  The
algorithm applies Centering theory to the problem of resolving
anaphoric reference.  Application of the algorithm requires three
basic steps.

\begin{enumerate}
\item {\sc generate} possible Cb--Cf combinations
\item {\sc filter} by constraints, e.g. contra-indexing, sortal predicates,
        centering  rules and constraints
\item {\sc rank} by transition orderings
\end{enumerate}

In order to apply this algorithm to Japanese, possible
Cb-Cf combinations, ({\sc generate} step 1), must be
constructed from the surface string and information from
the subcategorization frame of the verb. First the verb
subcategorization is examined, and if there are more
entities than appear in the surface string, zeros are
postulated as forward centers. These zeros are then treated
just like pronouns in English by the rest of the algorithm.
We use a different ranking for the Cf for Japanese
than for English, but this has no effect on the actual
algorithm itself since the Cf ranking is a declarative
parameter.

The steps of the algorithm can be interleaved to
improve computational efficiency \cite{BFP87}. Some simple
modifications are:
\begin{itemize}
\item Never propose a Cf that violates linguistic
constraints on contra-indexing. (In other words, apply the
contra-indexing filter as early as possible to avoid Cb-Cf
combinations that will be eliminated by that filter.)
\item If there are pronouns in an utterance, only
propose pronouns as possible Cbs. (Collect the pronouns
from the proposed Cfs as Cbs, from Rule 1)
\end{itemize}

In addition, it is simple to add additional filters to step
(2) of the algorithm.  For instance, any constraint that is
lexically specified such as [$\pm$animacy] can be easily
applied as a filter.  It is also possible to pursue a `best
first' strategy by interleaving steps (1), (2) and (3) so
that a {\sc continue} will be found without extra processing
if one exists.

In example \ex{1}, we illustrate in more detail how the steps
of the algorithm work and the difference between {\sc
continue} and {\sc retain}. Each utterance shows what the Cb
and Cf would be for that utterance. We will mostly be
concerned with the process of resolving the two zeros in
utterance \ex{1}c.

\eenumsentence{\item[a.]
\shortex{7}
{Taroo &wa &saisin &no &konpyuutaa &o &kaimasita.}
{ &{\sc  top/subj} &newest &of&computer &{\sc obj} &bought}
{{\it Taroo bought a new computer.}}

\begin{tabular}{|llllll|}
\hline
{\bf Cb:} & {\sc taroo} & & & & \\
{\bf Cf:} & [{\sc taroo}, {\sc computer}] & & & & \\ \hline
\end{tabular}

\item[b.]
\shortex{7}
{      0 &John &ni &sassoku &sore &o &misemasita.}
{{\sc subj}&John &{\sc obj2}&at once &that &{\sc obj} &showed }
{{\it (Taroo) showed it at once to John.}}

\begin{tabular}{|llllll|}
\hline
{\bf Cb:} & {\sc taroo} & & & & \\
{\bf Cf:} & [{\sc taroo}, {\sc john}, {\sc computer}] & & & & {\sc
continue} \\ \hline
\end{tabular}

\item[c.]
\shortex{7}
{      0 &0 &atarasiku &sonawatta &kinoo &o &setumeisimasita. }
{{\sc  subj} &{\sc obj2}&newly &equipped &function&{\sc  obj} &explained}
{{\it (Taroo) explained the newly equipped functions to (John).}}

\begin{tabular}{|llllll|}
\hline
{\bf Cb:} & {\sc taroo} &  & & &\\
{\bf Cf1:} & [{\sc taroo}, & {\sc john}] &  {\sc continue} & & 27  \\
& {\sc subj} & {\sc obj}&  & &\\  \hline
{\bf Cf2:} & [{\sc john}, & {\sc taroo}] &  {\sc retain} & & 1 \\
& {\sc subj}&  {\sc obj} &  & & \\  \hline
{\bf Cf3:} & [{\sc john}, & {\sc john} ] & {\sc contra-index filter} & &  \\
& {\sc subj} & {\sc obj} & & & \\  \hline
{\bf Cf4:} & [{\sc taroo}, & {\sc taroo}] & {\sc contra-index filter} & &  \\
& {\sc subj} & {\sc obj} & & & \\  \hline
\end{tabular}
\label{cont-ret-ex}
}

Example \ex{0}(c) has {\it explained} as the main verb, which requires
an animate subject and object2. Since there are two animate zeros in
\ex{0}c, which are also contra-indexed by syntactic constraints, both
Ziroo and Taroo must be realized in \ex{0}c. Constraint (3) restricts
the Cb to Taroo as
the highest ranked element from the Cf(\ex{0}b). The
interpretive process must also generate the possible candidates for
the Cf. If no constraints applied, then all 4 candidates shown
above as Cf1, Cf2, Cf3, and Cf4 would be possible.  However the
contraindexing filter will rule out Cf3 and Cf4.  As mentioned above,
there is no reason that these filters cannot be applied at the {\sc
generate} phase rather than later on.

The only {\sc continue} interpretation available, {\em Taroo explained
the newly equipped functions to John}, corresponds to the forward
centers Cf1.  It is a {\sc continue} interpretation because
Cb(\ex{0}c) = Cb(\ex{0}b) and also Cb(\ex{0}c) = Cp(\ex{0}c).  The
{\sc retain} interpretation is less preferred and is defined by the
fact that Cb(\ex{0}c) = Cb(\ex{0}b), but Cb(\ex{0}c) $\neq$
Cp(\ex{0}c).  This example supports the claim that a {\sc continue} is
preferred over a {\sc retain}($Z = 13.24, p < .001$).

In order to find this preferred continue interpretation in a `best
first' fashion, Taroo as the $\rm{Cp(U_{i-1})}$ would be tried first
as the $\rm{Cb(U_{i})}$, and as the interpretation for the subject.
Contraindexing rules out Taroo as the object, so John would be tried
next as the object.

In the next section, we examine further the application of
centering to the interpretation of zeros in
Japanese. We will examine the ranking of forward centers
that we have adopted for Japanese and explain how this is
partially determined by the way the Japanese language allows a
speaker to express discourse functions. We will also give some
examples of the interpretation of zeros in cases involving
Japanese discourse markers for {\sc topic} and {\sc empathy}.

\section{Centering in Japanese}
\label{jap-cent-sec}

The theory of centering is a formal specification that is
intended to model attentional state and is defined by the
rules and constraints given in section \ref{rules-sec}.
Attentional state in turn constrains the discourse
participant's interpretation process; one aspect of
attentional state is the notion of discourse salience.  In
the centering model, the ordering of the forward centers is
an approximation of discourse salience. This in turn is the
main determinant of discourse interpretation processes such
as the resolution of zeros in Japanese.  A crucial question
then is what discourse factors must be considered to
determine the ordering of the forward centers, $\rm{Cf}$,
in Japanese discourse.

Being a subject has been shown to be an important factor for English;
this is reflected in a Cf ordering by grammatical function
 \cite{Prince81,BFP87,Hudson88,Brennan93}. Aspects of surface order may
also affect the interpretation  \cite{Dieugenio90,Hajicova82}. An
interpretation algorithm can also use pronominalization as an
indicator of what the speaker believes is salient \cite{GJW86}.
Furthermore, zeros in Japanese are not realized syntactically so that
there must be a way to distinguish zeros from other entities inferred
to be part of a discourse situation. Consider:

\enumsentence{
\shortex{4}
{Taroo&ga&0&aimasita.}
{Taroo&{\sc subj}& {\sc obj2} &met}
{{\it Taroo met (0).}}
}

This sentence is not felicitous unless the addressee has already been
given some information about the person that Taroo met, either in the
current discourse or in previous discourses.  In contrast,
nonsubcategorized-for arguments like adjuncts are not necessarily
given a specific interpretation, but rather a non-specific one.

\enumsentence{
\shortex{7}
{Taroo&ga&Hanako&ni&aimasita.}
{Taroo&{\sc subj}&Hanako&{\sc obj2}&met}
{{\it Taroo met Hanako.}}
}

The sentence means that Taroo met Hanako at some time in some place:
the temporal-location of the {\it meeting\/} situation need not be
specified.  The speaker can utter this sentence even if the addressee
does not know where and when Taroo met Hanako.  Thus, in this work, we
only represent obligatorily subcategorized arguments of the verb on
the Cf, assuming that the salience of discourse entities is partially
determined by virtue of filling a verb's argument role, and the
information from the subcategorization frame is used to determine that
a zero is present in an utterance.

Zeros are then interpreted with reference to the current context.
Prince has proposed that the current context should be categorized by
{\sc assumed familiarity} \cite{Prince81,Horn86}, with a concomitant
goal of determining the correlation between the use of certain
linguistic forms and the types of assumed familiarity.  The first
division of assumed familiarity is into the subtypes of {\sc new},
{\sc inferable} and {\sc evoked}. {\sc new} can be divided into {\sc
brand-new}, discourse entities that are both new to the discourse and
new to the hearer, and {\sc unused}, discourse entities old to the
hearer but new to the discourse. The information status of {\sc
evoked} can be further divided into {\sc textually evoked}, old in the
discourse and therefore old to the hearer as well, and {\sc
situationally evoked}, entities in the current situation.  {\sc
inferables} are technically both hearer-new and discourse-new but
depend on information that is old to the hearer and the discourse, and
are often treated by speakers as though they were both hearer-old and
discourse-old. There is a hierarchy of assumed familiarity in terms of
discourse salience:

\begin{quote}
{\bf Assumed Familiarity Hierarchy} (Prince 1981): \\
{\sc textually evoked} $>$ {\sc situationally evoked} $>$ {\sc
inferable} $>$ {\sc unused} $>$ {\sc brand-new}
\end{quote}

Zeros typically refer to {\sc evoked} entities,\footnote{Under certain
circumstances that we cannot explore here, it appears that zeros can
at times be used to refer to inferable or unused entities, just as
pronouns in English sometimes can be.} but there is a scale of
relative salience among the {\sc evoked} entities. In our theory this
is modeled with Cf ranking. We repeat the proposed ranking of the Cf
here and justify it in the following sections:\footnote{This ranking
resembles Kuno's Empathy Hierarchy and Kameyama's Expected Center
Order, but we distinguish two kinds of {\sc topic} and we posit that
{\sc object2} is more salient than {\sc object}.  We continue Kuno's
use of the term {\sc empathy} to represent the {\sc empathy locus},
whereas Kameyama used the property {\sc ident} for {\sc
empathy} \cite{Kameyama86a}.}

\begin{quote}
{\bf Cf Ranking for Japanese} \\
{\sc (grammatical OR zero) topic} $>$ {\sc empathy} $>$ {\sc subject} $>$ {\sc
object2}
$>$ {\sc object} $>$ {\sc others}
\end{quote}

The relevance of the notions of {\sc topic} and speaker's {\sc
empathy} to centering is that a discourse entity realized as the {\sc
topic} or the {\sc empathy locus} is more salient and should be ranked
higher on the Cf. Whenever a discourse entity simultaneously fulfills
multiple roles, the entity is usually ranked according to the highest
ranked role.

In the following sections we will discuss the motivation for this
ranking. Section \ref{topic-sec} discusses the role of the grammatical
topic marker {\it wa} in Japanese.  Section \ref{emp-sec} explains
the role of {\sc empathy} in Japanese discourse salience and shows
that {\sc (grammatical OR zero) topic} $>$ {\sc empathy} and that {\sc
empathy} $>$ {\sc subj}.  Section \ref{emp-cont-ret-sec} shows how the
centering algorithm handles utterances with empathy loci. Zero topics
will not be discussed until section \ref{zta-sec}.

\subsection{Topic}
\label{topic-sec}

Discourse entities that are {\sc evoked}, {\sc inferable} or {\sc
unused} can be marked as the {\sc topic}. The speaker cannot mark an
entity as the grammatical {\sc topic} unless the hearer is aware of
the object that s/he is going to talk about  \cite{Prince78b,Kuno76a}.
For example:

\enumsentence{
 \shortex{8}{Hutari&wa&paatii&ni&kimasita.}
{two-person&{\sc top/subj}&party&to&came} {\it Speaking of
two persons, they came to the party.}}

Example \ex{0} is felicitous only when {\it hutari\/} (`two
persons') is understood as meaning {\it the two people under
discussion\/}. The sentence never means that the people who
came to the party numbered two.

The fact that the {\it wa}-marked entity should be discourse-old is
also shown by the fact that a wh-question cannot be answered with a
{\it wa}-marked {\sc np}.

\eenumsentence{
\item[a.]
\shortex{8}{Dono&hito&ga&Ziroo&o&bengosimasita&ka.}
           {which&person&{\sc subj}&Ziroo&{\sc obj}&defended&{\sc q}}
           {\it Which person defended Ziroo?}

\item[b-1.]
\shortex{7}{Taroo&ga&Ziroo&o&bengosimasita.}
           {Taroo&{\sc subj}&Ziroo&{\sc obj}&defended}
           {\it Taroo defended Ziroo.}
\item[b-2.]
\shortex{7}{*Taroo&wa&Ziroo&o&bengosimasita.}
           {\ Taroo&{\sc top/subj}&Ziroo&{\sc obj}&defended}
           {\ \it Taroo defended Ziroo.}}

What the question context shows is that even in a simple
declarative sentence, the use of the topic marker {\it wa}
contrasts with the subject marker {\it ga} in what is
understood as already in the discourse context.  For
instance, in a discourse initial utterance, \ex{1}a assumes
no shared information or that {\it someone defended Ziroo}
and asserts that the someone is Taroo. In
\ex{1}b, the discourse-old proposition is that {\it Taroo did something} and
what is
asserted is that what he did was to defend Ziroo.

\eenumsentence{
\item[a.]
\shortex{7}{Taroo&ga&Ziroo&o&bengosimasita.}
           {Taroo&{\sc subj}&Ziroo&{\sc obj}&defended}
           {\it Taroo defended Ziroo.}

\item[b.]
\shortex{7}{Taroo&wa&Ziroo&o&bengosimasita.}
           {Taroo&{\sc top/subj}&Ziroo&{\sc obj}&defended}
           {\it Taroo defended Ziroo.}}

While topics are often subjects, subject and grammatical
topic need not coincide.  Any argument can be realized as a
topic, as shown in examples \ex{1} and \ex{2}.

\enumsentence{
\shortex{5}{Taroo & wa& Hanako & ga & bengosita.}
{Taroo&{\sc top}&Hanako&{\sc subj} &defended}
{{\it As for Taroo, Hanako defended (him).}}
}

\enumsentence{
\shortex{6}
{Tokyoo& e & wa & Hanako & ga & itta.}
{Tokyo &to &{\sc top} &Hanako&{\sc subj} & went}
{{\it To Tokyo, Hanako went.}}
}

The assumption that the {\sc topic} is more salient than the
{\sc subject\/}, when the two are different, is supported by
the fact that an indefinite {\sc np} in subject position such
as {\it who\/}, {\it which\/}, or {\it somebody\/} cannot be
regarded as the {\sc topic}: an indefinite {\sc np} is never
marked by the topic marker {\it wa\/}, but by the subject
marker {\it ga\/}.  For example:

\enumsentence{
\shortex{8}{Dono&hito&ga&Ziroo&o&bengosimasita&ka.}
           {which&person&{\sc subj}&Ziroo&{\sc obj}&defended&{\sc q}}
           {\it Which person defended Ziroo?}}

\enumsentence{
\shortex{8}{*Dono&hito&wa&Ziroo&o&bengosimasita&ka.}
           {\ who&person&{\sc top/subj}&Ziroo&{\sc obj}&defended&{\sc q}}
           {\ \it Which person defended Ziroo?}}

It is clear from these examples that the grammatical topic, {\it
wa}-marked entity, in Japanese, represents assumable shared
information in an on-going conversation.  It has been taken to be the
`theme' or `what the sentence is about'  \cite{Kuno73,Shibatani90}. In
our framework, this is the role of the Cb.  We will provide evidence
supporting this position in section \ref{cb-instant-sec}.  However we
claim that this is just a default and that other factors can
contribute to establishing or continuing an entity as the Cb. Kuno
also claims that a zero subject is equivalent to a {\it wa}-marked
entity and we provide support for this claim in section
\ref{zta-sec}, showing that the property of having previously
been the Cb, in combination with being realized by a zero, contributes
to an entity being the Cp.

\subsection{Empathy}
\label{emp-sec}

Kuno (1976) proposed a notion of {\sc empathy} in order to present the
speaker's position or identification in describing a situation.  In a
{\it hugging\/} situation involving a man named {\it Taroo\/} and his
son {\it Saburoo\/}, Kuno notes that this situation can be described
in various ways, some of which are shown in \ex{1}.

\eenumsentence{
\item[a.] Taroo hugged Saburoo.

\item[b.] Taroo hugged his son.

\item[c.] Saburoo's father hugged him.
}

These sentences differ from each other with respect to {\it camera
angle\/}, the position that the speaker takes to observe and describe
this situation. In \ex{0}a, the speaker is assumed to be describing
the event objectively: the camera is placed at the same distance from
both {\it Taroo\/} and {\it Saburoo\/}. On the other hand, the camera
may be placed closer to {\it Taroo\/} in \ex{0}b and closer to {\it
Saburoo\/} in \ex{0}c.  This is shown by the use of relational terms
such as {\it son\/} and {\it father\/}, respectively.  The term {\sc
empathy\/} is used for this {\it camera angle\/}, which indicates the
speaker's position among the participants in the event
described.\footnote{The speaker's position is not determined by his
physical proximity but also measured by the emotional or social
relationship. In this sense, the term {\it speaker's
identification\/} \cite{Kuno76a} may be more suitable than the term {\it
speaker's position\/}.  Furthermore, the notion of {\sc empathy} is
different from that of perspective  \cite{Iida92}. Empathy is the
speaker's identification with a discourse entity, but the speaker does
not have to take the perspective of the person who he empathizes with.
For example, consider the following utterance:

\shortex{10}
{(i)&Taroo & wa & Hanako & ni & migigawa & no & hon & o
& totte-kureta.}
{    &Taroo& {\sc top/subj} &Hanako&{\sc obj2} & right &GEN & book
&{\sc obj}&take-gave }
{\ \ \ {\it Taroo did Hanako a favor in taking a book
on his/her right.}}

In this example, the speaker empathizes with Hanako as
indicated by the empathy verb {\it kureru\/}, yet he still
can describe the given situation from Taroo's perspective,
which is indicated by ambiguity in the interpretation of
the deictic expression {\it migigawa no\/} (`right of').
}

In Japanese the realization of speaker's empathy is
especially important when describing an event involving
{\it giving\/} or {\it receiving\/}. There
is no way to describe a {\em giving\/} and {\em
receiving\/} situation objectively  \cite{Kuno-Kab77}.  In
\ex{1}, the use of the verb {\em kureru\/} indicates the
speaker's empathy with Ziroo, the discourse entity realized
in object position, while in \ex{2}, the speaker's
empathy with the subject Taroo is indicated by the
use of the past tense form {\it yatta} of the verb {\em
yaru\/}.

\enumsentence{
\shortex{7}{Taroo&ga&Ziroo&ni&hon&o&kureta.}
           {Taroo&{\sc subj}&Ziroo&{\sc obj2}&book&{\sc obj}&gave}
           {{\it Taroo gave Ziroo a book.}\ \ \ \ \ \
            {\sc empathy=obj2=ziroo}}}

\enumsentence{
\shortex{7}{Taroo&ga&Ziroo&ni&hon&o&yatta.}
           {Taroo&{\sc subj}&Ziroo&{\sc obj2}&book&{\sc obj}&gave}
           {{\it Taroo gave Ziroo a book.}\ \ \ \ \ \
            {\sc empathy=sub=taroo}}}

A verb that is sensitive to the speaker's
empathy is an {\sc empathy-loaded} verb. The {\sc
empathy locus} is the argument position whose referent the
speaker automatically identifies with. In other words, the
verb {\it kureru\/} has the {\sc empathy locus\/} on the
object, while verbs like {\it yaru\/} place the {\sc empathy
locus} on the subject.

The use of deictic verbs such as {\it kuru\/} (`come'), {\it
iku\/} (`go'), {\it okuru\/} (`send to'), and {\it yokosu\/}
(`send in') also encode speaker's empathy.  For
example, the speaker indicates empathy
with Taroo by using the past tense form {\it kita}
of the verb {\it kuru} in the following example.

\enumsentence{\shortex{7}{Hanako&wa&Taroo&no&tokoro&ni&kita.}
         {Hanako&{\sc top/subj}&Taroo&of&place&to&came}
         {\it Hanako came to Taroo's place.}
          }

Many Japanese verbs can be made into empathy-loaded verbs
due to a productive verb-compounding operation by which these
empathy-loaded verbs are used as the auxiliary verb,
attaching to the main verb.\footnote{Certain intransitive verbs
cannot be made into empathy-loaded verbs since the
empathy-loaded versions make no sense, e.g.
{\it moreru} (leak).}  For example, {\em kureru}
can be used as a suffix, to mark {\sc obj} or {\sc obj2} as
the {\sc empathy locus}.  The attachment of {\em yaru} marks
{\sc subject} as the {\sc empathy locus}.  The complex
predicate made by this operation inherits the {\sc empathy
locus} of the suffixed verb.  For example:

\enumsentence{
\shortex{7}
{Hanako&ga&Taroo&ni&hon&o&yonde-kureta.}
{Hanako&{\sc subj}&Taroo&{\sc obj2}&book&{\sc obj}&read-gave}
{\it Hanako did Taroo a favor in reading a book.\ \ \ {\sc empathy = obj2 =
taroo}}
}

In this case Taroo is interpreted as the {\sc empathy
locus} due to the auxiliary {\it kureta} attached to the
main verb.  Similarly in \ex{1}, the speaker indicates
empathy with Hanako by using the past tense form {\it yatta}
of the verb {\it yaru} as an auxiliary verb to the main verb
{\it tazuneru}.

\enumsentence{
\shortex{7}
{Hanako&ga&Taroo&o&tazunete-yatta.}
{Hanako&{\sc subj}&Taroo&{\sc obj}&visit-gave}
{(lit.)\it Hanako received a favor in visiting Taroo.
\ \ \  {\sc empathy = subj = hanako}}  }

As demonstrated in the following examples, a discourse entity
that is realized as the {\sc empathy locus} must be {\sc evoked}.

\enumsentence{
\shortex{7}{Taroo&ga&Ziroo&ni&okane&o&kasite-kureta.}
           {Taroo&{\sc subj}&Ziroo&{\sc obj2}&money&{\sc obj}&lend-gave}
           {\it Taroo did Ziroo a favor in lending him some money.}}

\enumsentence{
\shortex{7}{*Taroo&ga&dareka&ni&okane&o&kasite-kureta.}
           {\ Taroo&{\sc subj}&somebody&{\sc obj2}&money&{\sc obj}&lend-gave}
           {\ \it Taroo did somebody a favor in lending him some money.}}

\enumsentence{
\shortex{10}{*Taroo&ga&misiranu&hito&ni&okane&o&kasite-kureta.}
           {\ Taroo&{\sc subj}&unknown&person&{\sc obj2}&money&{\sc
obj}&lend-gave}
           {\ \it Taroo did a stranger a favor in lending him some money.}}

The contrast between \ex{-2}, \ex{-1}, and \ex{0} demonstrates that
the use of a {\sc brand-new} entity in the {\sc empathy locus}
position of the verb {\it give\/} is not acceptable.
Therefore
an entity in the {\sc empathy}-locus position
is ranked in a higher position on the Cf than the subject.

\subsubsection{Empathy and the Centering Algorithm}
\label{emp-cont-ret-sec} Using the Centering Algorithm, we model {\sc
empathy} as a language-specific discourse factor by adding the {\sc
empathy}-marked discourse entity to the Cf ranking. Then preferences
for {\sc continue} over {\sc retain} when {\sc empathy} is involved
can be demonstrated, as in example \ex{1} below:\footnote{The verb
form {\it kuremasita\/} in (\ex{1})b is the polite form of {\it
kureta}, the past tense form of the verb {\it kureru}.}

\eenumsentence{
\item[a.]
\shortex{8}{Hanako&wa&kuruma&ga &kowarete& komatteimasita.}
         {Hanako&{\sc top/subj}&car&{\sc subj}&broken&at a loss-was}
         {\it Her car broken, Hanako was at a loss.}

\begin{tabular}{|llllll|}
\hline
{\bf Cb:} & {\sc hanako} & & & & \\
{\bf Cf:} & [{\sc hanako}, {\sc car}] & & & & \\ \hline
\end{tabular}

\item[b.]
\shortex{8}{Taroo&ga& 0  &sinsetu-ni &te&o&kasite-kuremasita.}
         {Taroo&{\sc subj}&{\sc obj2/emp}&kindly&hand&{\sc obj}&lend-gave.}
         {\it Taroo kindly did (Hanako) a favor in helping her.}

\begin{tabular}{|llllll|}
\hline
{\bf Cb:} & [{\sc hanako}] & & & & \\
{\bf Cf:} & [{\sc hanako}, & {\sc taroo}]  & & & \\
    & {\sc empathy} & {\sc subj} & & & \\ \hline
\end{tabular}

\item[c.]
\shortex{8}{Tugi&no&hi&0&0&eiga&ni&sasoimasita.}
         {next&of&day&{\sc subj}&{\sc obj}&movie&to&invited}
         {\it Next day (Hanako) invited (Taroo) to a movie.}

\begin{tabular}{|llllll|}
\hline
{\bf Cb:} & {\sc hanako} & & & & \\
{\bf Cf1:} & [{\sc hanako}, & {\sc taroo}] & {\sc continue} & & 16 \\
& {\sc subj} & {\sc obj} & & & \\  \hline
{\bf Cf2:} & [{\sc taroo}, & {\sc hanako} ] & {\sc retain} & & 2 \\
& {\sc subj}  & {\sc obj} & & & \\  \hline
\end{tabular}
\label{emp-cont-ret-ex}
}

In \ex{0}c, the verb {\it invited} requires an animate subject and
object, and these must be realized by different discourse entities due
to the contraindexing constraint. Hanako is the most highly ranked
entity from \ex{0}b that is realized in \ex{0}c, and therefore must be
the $\rm{Cb}$. The preferred interpretation is therefore {\em she
invited him to a movie} ($Z = 5.25, p < .001$). This corresponds to
Cf1, the more highly ranked {\sc continue} transition, in which Hanako
is the preferred center, $\rm{Cp}$. This interpretation can be found
with minimal processing by trying the Cp(\ex{0}b), Hanako, as the
Cb(\ex{0}c), by interpreting the subject zero as Hanako. This gives a
{\sc continue} transition. Then contraindexing constraints mean that
Hanako cannot fill both argument positions, so the object position is
interpreted as Taroo. This interpretation is found with minimal
processing by interleaving the steps of the centering algorithm
proposed in  \cite{BFP87}.

Note that nothing special needs to be said about the fact that {\sc
empathy} is the discourse factor that made Hanako the Cp in \ex{0}b
and thus predicted that Hanako would be the Cb at \ex{0}c ({\it pace}
 \cite{BFP87}).  The preference in the interpretation follows from the
distinction between {\sc continue} and {\sc retain} and the ranking of
$\rm{Cf}$.  Thus, the centering framework is easily adapted to handle
this language specific feature.

\subsection{Topic and empathy}

In general the assignment of the {\sc empathy} relationship is
pragmatic.  It is determined by the speaker's relation to the
discourse participants in the discourse.  In \ex{0}, for
example, the {\sc empathy} relationship between the speaker and {\it
Hanako\/} and between the speaker and {\it Taroo\/} is clear: the use
of the empathy verb in the second sentence indicates that the speaker
is closer to {\it Hanako\/} than to {\it Taroo\/}.

However, besides cases where the speaker clearly expresses who s/he empathizes
with, it is also possible for the context to provide some information about the
speaker's proximity relationship with discourse participants in the given
discourse, so that the hearer can determine the {\sc empathy} relation that the
speaker has in mind.  In this paper, we only consider cases where {\sc empathy}
is syntactically marked by the use of empathy-loaded verbs.

Kuno's notion of {\sc empathy} is more general.  For instance, Kuno's
{\sc empathy hierarchy\/} consists of different scales for {\sc
empathy} that include notions such as {\sc topic} and {\sc speaker}
 \cite{Kuno87}. Kuno's Topic Empathy Hierarchy suggests that the
discourse entity realized as the {\sc topic} will often coincide with
the {\sc empathy locus}:

\begin{quote}
{\it Topic Empathy Hierarchy: Discourse-Topic $>$
Discourse-Nontopic} \\ Given an event or state that
involves A and B such that A is coreferential with the
topic of the present discourse and B is not, it is
easier for the speaker to empathize with A than with B
\end{quote}

In support of Kuno's claim, we have found that when no empathy
relation is clearly indicated and no topic has been clearly
established that it is difficult for a hearer to determine the empathy
relation that the speaker intends. Previous Cbs and current Cps can be
high on the empathy scale, and yet the discourse entity realized as
the grammatical {\sc topic} does not necessarily coincide with the
discourse entity realized as the {\sc empathy locus}. A simple
sentence to show this point is given below:

\enumsentence{
\shortex{7}{Taroo&wa&Ziroo&ni&hon&o&yonde-kuremasita.}
         {Taroo&{\sc top/subj}&Ziroo&{\sc obj2}&book&{\sc obj}&read-gave}
         {{\it Taroo gave Ziroo a favor of reading a book.}\ \ \ {\sc empathy =
obj2 = ziroo}}
}

In example \ex{0}, Taroo is the {\sc topic} while Ziroo is the
{\sc empathy locus}.  Similarly, a zero does not have to be
realized as the {\sc empathy locus}. In \ex{1}b the zero in
subject position realizes the Cb and refers to Taroo.

\eenumsentence{\item[a.]
\shortex{7}{Taroo&wa&syukudai&o&zenbu&yari-oemasita.}
         {Taroo&{\sc top/sub}&homework&{\sc obj}&all&do-finished}
         {\it Taroo finished his homework.}

\item[b.]
\shortex{7}{0&Ziroo&ni&hon&o&yonde-kuremasita.}
         {{\sc subj}&Ziroo&{\sc obj2}&book&{\sc obj}&read-gave}
         {\it (Taroo) gave Ziroo a favor of reading a book.\ \ \ {\sc empathy =
obj2 = ziroo}}
}

{\sc topic} is higher than {\sc empathy} in the Cf ranking.  The
higher degree of salience of {\sc topic} over {\sc empathy} is shown
by the different interpretation of (b) sentences in examples \ex{1}
and \ex{2}. The only difference in these examples is that Mitiko is
{\it wa}-marked in \ex{1}a but is {\it ga}-marked in \ex{2}a:

\eenumsentence{\item[a.]
\shortex{8}{Mitiko&wa&kanai&o&gityoo&ni&osite-kuremasita.}
           {Mitiko&{\sc top/subj}&wife&{\sc obj/emp}&chairman&{\sc
obj2}&recommend-gave}
{\it Mitiko did my wife a favor in recommending her as chairperson.}

\item[b.]
\shortex{10}{0\ &asu&no&kaihyoo-kekka&o&tanosimi-ni&siteim asu.}
           {{\sc subj}&tomorrow&of&results&{\sc obj}&look-forward&doing-is}
           {\it (Mitiko) is looking forward to tomorrow's results.}
}

\eenumsentence{\item[a.]
\shortex{8}{Mitiko&ga&kanai&o&gityoo&ni&osite-kuremasita.}
           {Mitiko&{\sc subj}&wife&{\sc obj/emp}&chairman&{\sc
obj2}&recommend-gave}
           {\it Mitiko did my wife a favor in recommending her as chairperson.}

\item[b.]
\shortexdt{10}{0\ &asu&no&kaihyoo-kekka&o&tanosimi-ni&siteimasu.}
           {{\sc subj}&tomorrow&of&results&{\sc obj}&look-forward&doing-is}
           {\it (Mitiko) is looking forward to tomorrow's results.}
           {\it (My wife) is looking forward to tomorrow's results.}
}

The {\sc topic} {\it Mitiko\/} is preferred as the unexpressed subject
of the (b) sentence in \ex{-1}. \footnote{The zero may be interpreted
as indirectly referring to the speaker. This interpretation is always
possible when the verb {\it kureru\/} is used: the use of {\it
kureru\/} implies that the speaker is closer to the beneficiary
argument (i.e. the {\it o\/}-marked {\sc np} in these examples), and
the favor given to this person is understood as a benefit to the
speaker as well.} On the other hand, the subject {\it Mitiko\/} is not
strongly preferred as shown in \ex{0}: the zero in the second sentence
in \ex{0} is understood as referring to either {\it Mitiko\/} or {\it
my wife\/}.  That is, the possible interpretation in these examples
shows that the {\sc np} {\it my wife}, which is realized as the {\sc
empathy locus}, is not as salient as the {\sc
topic}.\footnote{Although it seems as though empathy isn't higher than
subject, the conflating factor is that topic marking establishes a Cb
whereas in \ex{0} no Cb has been established.  This is explained in
detail in section \ref{cb-instant-sec}.}

So why is it easier to empathize with a discourse entity
that has been the topic as Kuno demonstrates?  It seems
important to keep the notions of {\sc topic} and {\sc
empathy} separate, but in section
\ref{emp-and-zta-sec} we will demonstrate an effect where
the topic entity is interpreted as the empathy locus.  We claim that
the ranking of the Cf and the potential for a {\sc continue}
interpretation determines whether this effect will hold.  In other
words, the tendency for the topic entity to be interpreted as the
empathy locus follows from more general discourse processing factors,
such as a hearer preferring {\sc continue} transitions
within a given local stretch of discourse.

\subsection{Summary}

To summarize, we have outlined the roles of discourse
markers such as those for {\sc topic} and {\sc empathy} by
which Japanese grammaticizes some aspects of discourse
function, and we have argued that {\sc topic} and {\sc
empathy} markers can only be used on entities that are
already in the discourse context.

One factor that hasn't been discussed is the role of
pronominalization, but many researchers have argued that
discourse entities realized by pronouns are more salient than
other discourse entities  \cite{CH77,GJW86,Kuno76a,Kuno87}. We
take zeros in Japanese to be analogous to pronouns in English
in this respect.  Since pronominalization can apply at any
position in the ranking of the Cf, the role of its
contribution is particularly interesting when it is in
conflict with some other factor such as grammatical function
or topic marking. This will be discussed further in section
\ref{zta-sec}.

\section{Initial Center Instantiation}
\label{cb-instant-sec}

{\sc Initial center instantiation} is a process by
which a discourse entity introduced in a
segment-initial utterance becomes the Cb. In our
framework, this happens as a side effect of the
Centering Algorithm.  Typically, when an interpretation
is found for the second utterance in a discourse
segment, the Cb
becomes instantiated.\footnote{In  \cite{WIC90} we called
this Center Establishment. Henceforth we will refer to
this process as Center Instantiation in order to avoid
confusion with Kameyama's term center establishment,
which is a different mechanism in her
theory  \cite{Kameyama85}.}  The $\rm{Cb}$ of an initial
utterance $\rm{U_{i}}$ is treated as a variable which
is then unified with whatever $\rm{Cb}$ is assigned to
the subsequent utterance $\rm{U_{i+1}}$.

Typically, a discourse entity is introduced as a {\it ga}-marked
subject, and then is referred to by a zero in a subsequent
utterance \cite{ClancyDowning87}.  Consider example \ex{1}.

\eenumsentence{
\item[a.]
\shortex{8}
{Taroo &ga &deeta &o &konpyuutaa &ni &utikondeimasita.}
{Taroo&{\sc subj}&data  &{\sc obj} &computer&in&was-storing}
{{\it Taroo was storing the data in a computer.}}

\begin{tabular}{|ll|}
\hline
{\bf Cb:} & [?]  \\
{\bf Cf:} & [{\sc taroo}, {\sc data}] \\ \hline
\end{tabular}

\item[b.]
\shortex{5}
{      0&yatto &hanbun &yari-owarimasita.}
{ {\sc subj}&finally &half &do-finished}
{{\it Finally (Taroo) was half finished.}}

\begin{tabular}{|lll|}
\hline
{\bf Cb:} & {\sc taroo} & \\
{\bf Cf:} & [{\sc taroo}] & {\sc continue} \\ \hline
\end{tabular}
}

Using Taroo as the subject in \ex{0}a is not enough to establish this
discourse segment as being about Taroo.  It is the use of the zero in
\ex{0}b that serves to instantiate Taroo as the Cb. By our definition
of {\sc continue}, \ex{0}b is a continue transition, because
Cb(\ex{0}b) $=$ Cp(\ex{0}b) and there was no Cb in \ex{0}a.  However,
Kuno argues that referring to a discourse entity with a zero is
equivalent to marking it as the grammatical topic with {\it wa}
 \cite{Kuno72}.  Our interpretation of this argument is that the use of
{\it wa} in a discourse initial utterance instantiates the {\it
wa}-marked entity as the Cb in one utterance.  This claim is supported
by the contrast with the {\sc ga}-{\sc wa} alternation in examples
\ex{1} and \ex{2}, where there is a shift in interpretation depending
on whether {\it Taroo\/} is marked with {\it wa} in the first
sentence.\footnote{These examples were tested by asking survey
participants to indicate preference rankings.  The numbers given here
are only for those subjects who expressed strong preferences; some
subjects expressed no preference.}

\eenumsentence{
\item[a.]
\shortex{9}
{Taroo &ga &Ziroo & o &min'na &no & mae &de & tatakimasita.}
{ &{\sc subj} & & {\sc obj} & all & of & front &in & hit.}
{{\it Taroo hit Ziroo in front of all the other people.}}

\begin{tabular}{|lll|}
\hline
{\bf Cb:} & [?] & \\
{\bf Cf:} & [{\sc taroo}, {\sc ziroo}] & \\ \hline
\end{tabular}

\item[b.]
\shortex{5}
{Itiniti-zyuu, & kanzen-ni & 0 &   0 &  musi-simasita.}
{   all-day &       completely & & &      ignored}
{{\it  (Ziroo) ignored (Taroo) all day.}}

\begin{tabular}{|lll|}
\hline
{\bf Cb:} & {\sc taroo} & \\
{\bf Cf:} & [{\sc taroo}, {\sc ziroo}] & 3 \\ \hline
\end{tabular}

\begin{tabular}{|lll|}
\hline
{\bf Cb:} & {\sc ziroo} & \\
{\bf Cf:} & [{\sc ziroo}, {\sc taroo}] & 8 \\ \hline
\end{tabular}
}

In example \ex{0}, {\it Taroo\/} is introduced by {\it ga}.
In this case, it appears that there is  tendency due to lexical
semantics to instantiate {\it
Ziroo} as the Cb in the second utterance.\footnote{The
number of subjects here are too small to test statistically.}

By the centering definitions, taking either Taroo
or Ziroo to be the Cb can result in a {\sc continue}
interpretation.  However, assuming that the Cf ordering at
\ex{0}a is correct, constraint 3 is violated by the
preferred interpretation of \ex{0}b.  Since both of the
entities in Cf(\ex{0}a) are realized, the Cb in \ex{0}b
should be the most highly ranked one.  There are two
possible conclusions here: (1) In discourse initial
utterances, when no clear indication of topic is given, the
Cf ordering alone is not a strong constraint; (2) the
ordering of the Cf should be partly determined by lexical
semantics or other knowledge about the situation being
described. However compare \ex{0} with \ex{1}.

\eenumsentence{
\item[a.]
\shortex{9}
{Taroo &wa &Ziroo & o &min'na &no & mae &de & tatakimasita.}
{ &{\sc subj} & & {\sc obj} & all & of & front &in & hit.}
{\it Taroo hit Ziroo in front of all the other people.}

\begin{tabular}{|lll|}
\hline
{\bf Cb:} & [{\sc taroo}] & \\
{\bf Cf:} & [{\sc taroo}, {\sc ziroo}] & \\ \hline
\end{tabular}

\item[b.]
\shortex{5}
{Itiniti-zyuu, & kanzen-ni & 0 &   0 &  musi-simasita.}
{   all-day &       completely & & &      ignored}
{\it (Taroo) ignored (Ziroo) all day.}

\begin{tabular}{|lll|}
\hline
{\bf Cb:} & {\sc taroo} & \\
{\bf Cf:} & [{\sc taroo}, {\sc ziroo}] & 10 \\ \hline
\end{tabular}

\begin{tabular}{|lll|}
\hline
{\bf Cb:} & {\sc ziroo} & \\
{\bf Cf:} & [{\sc ziroo}, {\sc taroo}] & 4 \\ \hline
\end{tabular}
}

The use of {\it wa} in \ex{0} seems to override the semantic
preference that was exhibited in \ex{-1}, so that subjects now prefer
an interpretation in which Taroo is the Cb.\footnote{The small number
of subjects means that we can't provide statistical support for this
claim.} This shows that Taroo has not been instantiated as the Cb when
it is time to interpret the two zeros in \ex{-1}b. We explain the
contrast by assuming that the {\sc topic} instantiates the Cb when it
is first introduced in a discourse initial utterance such as in
\ex{0}a. Then the only way to get a {\sc continue} interpretation for
\ex{0}b is for Taroo to be the Cb at \ex{0}b.

Furthermore, we can detect no differences in the interpretation of the
final utterance between 3 utterance sequences in which an entity is
introduced by {\it wa}, and 4 utterance sequences in which an entity
is first introduced by {\it ga} and then realized by a zero in the
second utterance. This provides further support for the claim that the
status of discourse entities realized as grammatical topics and those
realized as zero subjects is equivalent.

\subsection{Summary}

In sum, we have argued that the use of {\it wa} in a discourse initial
utterance instantiates the {\it wa}-marked entity as the Cb.  Cb
instantiation can equivalently be done with a 2 utterance sequence in
which the entity is first introduced as a subject, {\it ga}-marked,
and then established as the Cb in the following utterance with a zero
referring to that entity.  In addition, the fact that the Cb is
uninstantiated in discourse initial utterances has the effect that the
Cf ranking in a discourse initial utterance is not a strong constraint
as it is once a Cb is established.

\section{Zero Topic Assignment}
\label{zta-sec}

In this section we introduce the notion of a {\sc zero
topic} and a rule or assumption that can be employed as part
of the interpretive process called {\sc zero topic
assignment}.

The rule of {\sc zero topic assignment} defines our distinction
between grammatical topic and zero topic.  This rule allows a zero
that has just been the Cb to continue as the Cp, even when it is not
realized in a discourse salient syntactic position such as subject.
We will demonstrate this with examples that realize both grammatical
and zero topics. In these cases, the discourse situation is such that
the hearer may maintain multiple hypotheses about where the speaker's
attention is directed, and must determine whether to apply the default
that the grammatical topic is usually the Cp.\footnote{While some of
the utterance sequences we examine are potentially ambiguous for
native speakers, the examination of these discourse situations offers
considerable insight into those where there is no ambiguity.}

\begin{quote}
{\bf Zero Topic Assignment} \\ When a zero in $\rm{U_{i+1}}$
represents an entity that was the $\rm{Cb(U_{i})}$,
and when no other {\sc continue}
transition is available, that zero
may be interpreted as the {\sc zero topic} of
$\rm{U_{i+1}}$.
\end{quote}

What this means is that, in certain discourse environments, the entity
that was previously the Cb is predicted to continue as the Cb.  We
conjecture that ZTA is applicable in all free word-order languages
with zeros.\footnote{We only look at object topics here but there may
be limits as to how lowly ranked on the Cf and entity can be and still
be the zero topic, e.g.  by-passive agentive.} However {\sc zero topic
assignment} is {\bf optional}; here we have suggested 2 constraints on
when it applies. We will give examples below of cases where it doesn't
apply.

The option of {\sc zero topic assignment} (henceforth ZTA)
has been overlooked in previous treatments of zeros in
Japanese. ZTA explains why the discourse entity Hanako, which
is realized as {\sc object2} in \ex{1}c is interpreted as
the {\sc subject} of \ex{1}d.

\eenumsentence{\item[a.]
\shortex{8}{Hanako &wa &siken &o &oete,&kyoositu&ni&modorimasita.}
          {Hanako&{\sc top/subj} &exam &{\sc obj} &finish &classroom &to
&returned}
          {{\it Hanako returned to the classroom, finishing her exams.}}

\begin{tabular}{|lll|}
\hline
{\bf Cb:} & {\sc hanako} & \\
{\bf Cf:} & [{\sc hanako}, {\sc exam}]   & \\ \hline
\end{tabular}

\item[b.]
\shortex{6}{0  &hon &o &locker &ni &simaimasita.}
           {{\sc subj} &book &{\sc obj} &locker &in &took-away}
           {{\it She put her books in the locker.}}

\begin{tabular}{|lll|}
\hline
{\bf Cb:} & {\sc hanako} & \\
{\bf Cf:} & [{\sc hanako}, {\sc book}] & {\sc continue} \\ \hline
\end{tabular}

\item[c.]
\shortex{10}{Itumo &no &yooni &Mitiko &ga &0&mondai no&tokikata&o
 &setumeisidasimasita.}
{always& &like &{\sc subj} &Mitiko &{\sc obj2} & problem&solve-way &{\sc obj}
&explained}
{{\it Mitiko, as usual,  explained (to Hanako) how to solve the problems.}}

\begin{tabular}{|lll|}
\hline
{\bf Cb:} & {\sc hanako} & \\
{\bf Cf1:} & [{\sc hanako}, {\sc mitiko}, {\sc solution}]  {\sc zta
continue} & \\
           & {\sc top}, {\sc subj}, {\sc obj} & \\  \hline
{\bf Cf2:} & [{\sc mitiko}, {\sc hanako}, {\sc solution}]  {\sc retain} & \\
           & {\sc subj}, {\sc obj2}, {\sc obj}   & \\  \hline
\end{tabular}

\item[d.]
\shortex{5}
{      0 &   0 &ohiru &ni &sasoimasita.}
{ {\sc subj} & {\sc obj} &lunch &to &invited }
{{\it (Hanako) invited (Mitiko) to lunch.}}

\begin{tabular}{|lll|}
\hline
{\bf Cb1:} & {\sc hanako} & \\
{\bf Cf1:} & [{\sc hanako},  {\sc lunch}, {\sc mitiko}]  {\sc continue}
from Cf1(c)& 28 \\
           & {\sc subj}, {\sc obj2}, {\sc obj} & \\  \hline \hline
{\bf Cb2:} & {\sc mitiko} & \\
{\bf Cf2:} & [{\sc mitiko}, {\sc lunch}, {\sc hanako}]  {\sc smooth-shift}
from Cf2(c)& 6\\
           & {\sc subj}, {\sc obj2}, {\sc obj}   & \\  \hline
\end{tabular}
\label{zta-ex-ga}
}

The possibility of ambiguity as to the attentional state of the
speaker is reflected in the fact that there are two possible Cfs for
\ex{0}c; Cf2 of \ex{0}c is the only Cf possible without ZTA, and
represents a {\sc retain} rather than a {\sc continue}.  By the
formulation of the ZTA rule above, ZTA is triggered by the fact that
no {\sc continue} transition is available.

The availability of ZTA means that {\sc hanako} can be
the Cp even when {\sc mitiko} is realized as the
subject.  This leads to a potential ambiguity in \ex{0}d, because it
is possible for a hearer to simultaneously entertain both of the
Cf(\ex{0}c).  In this case the ZTA interpretation is preferred ($ Z =
4.95, p < .001$). The less preferred {\sc smooth-shift} interpretation
would result from the algorithm's application to Cf2 of
\ex{0}c.\footnote{ See section
\ref{cent-sec} for an example of how a smooth-shift interpretation is
calculated.}

ZTA explains the contrast between the discourse segments in
example \ex{0} above and \ex{1} below.  The only difference
between \ex{0} and \ex{1} is that in \ex{0}c, {\sc mitiko} is
a {\it ga}-marked subject, whereas in \ex{1}c, {\sc mitiko}
is a {\it wa}-marked subject/grammatical topic.  Utterances \ex{0}c and
\ex{1}c have the same meaning.  This minimal pair provides a
test to see whether ZTA actually characterizes these
discourse related effects.

\eenumsentence{
\item[a.]
\shortex{8}{Hanako &wa &siken &o &oete,&kyoositu&ni&modorimasita.}
          {Hanako&{\sc top/subj} &exam &{\sc obj} &finish &classroom &to
&returned}
          {{\it Hanako returned to the classroom, finishing her exams.}}

\begin{tabular}{|lll|}
\hline
{\bf Cb:} & {\sc hanako} & \\
{\bf Cf:} & [{\sc hanako}, {\sc exam}]   & \\ \hline
\end{tabular}

\item[b.]
\shortex{6}{0  &hon &o &locker &ni &simaimasita.}
           {{\sc subj} &book &{\sc obj} &locker &in &took-away}
           {{\it (Hanako) put (her) books in the locker.}}

\begin{tabular}{|lll|}
\hline
{\bf Cb:} & {\sc hanako} & \\
{\bf Cf:} & [{\sc hanako}, {\sc book}]  {\sc continue} & \\ \hline
\end{tabular}

\item[c.]
\shortex{10}{Itumo &no &yooni &Mitiko &wa &0&mondai no&tokikata&o
 &setumeisidasimasita.}
{always& &like &{\sc top/subj} &Mitiko &{\sc obj2} & problem&solve-way &{\sc
obj}
&explained}
{{\it Mitiko, as usual,  started explaining (to Hanako) how to solve the
problems.}}

\begin{tabular}{|lll|}
\hline
{\bf Cb:} & {\sc hanako} & \\
{\bf Cf1:} & [{\sc hanako}, {\sc mitiko}, {\sc solution}]  {\sc zta
continue} & \\
           & {\sc top}, {\sc subj}, {\sc obj} & \\  \hline
{\bf Cf2:} & [{\sc mitiko}, {\sc hanako}, {\sc solution}]  {\sc retain} & \\
           & {\sc top}, {\sc obj2}, {\sc obj}   & \\  \hline
\end{tabular}

\item[d.]
\shortexdt{5}
{      0 &   0 &ohiru &ni &sasoimasita.}
{ {\sc subj} & {\sc obj} & lunch &to &invited }
{{\it (Hanako) invited (Mitiko) to lunch.}}
{{\it (Mitiko) invited (Hanako) to lunch.}}

\begin{tabular}{|lll|}
\hline
{\bf Cb1:} & {\sc hanako} & \\
{\bf Cf2:} & [{\sc hanako}, {\sc lunch}, {\sc mitiko}]  {\sc continue}
from Cf1(c)& 18 \\
           & {\sc subj}, {\sc obj2}, {\sc obj} & \\  \hline
\hline
{\bf Cb2:} & {\sc mitiko} & \\
{\bf Cf2:} & [{\sc mitiko}, {\sc lunch}, {\sc hanako} ]  {\sc smooth-shift}
from Cf2(c)& 16 \\
           & {\sc subj}, {\sc obj2}, {\sc obj}   & \\  \hline
\end{tabular}
\label{zta-ex-wa}
}

The {\it wa} marking has the predicted effect.  Using
the grammatical topic marker {\it wa} in \ex{0}c
dampens ZTA and thus affects the interpretation of
\ex{0}d, which is now completely ambiguous ($Z = 0.34$,
not significantly different than chance).
Because the discourse entity realized as the grammatical topic and
indicated by the {\it wa}-marked {\sc np} is the Cp by
default, 10 subjects who previously did, can no longer get an
interpretation that depends on ZTA.  The situation
can be characterized as a case of competing defaults, so
that in \ex{0}, some
hearers apply the default that the {\it wa}-marked entity is the
Cp, and others apply the default that {\sc continue} interpretations
are preferred and that zeros realize discourse entities that are
ranked highly on the Cf.

The {\sc retain} interpretation in \ex{0}c, Cf2, indicates that these
hearers expect the conversation to shift to being about Mitiko; the
fact that Mitiko is the Cp(\ex{0}c), along with constraint 3 will
force a shift.  Given a {\sc shift}, the {\em Mitiko invited Hanako to
lunch} interpretation is preferred because it is the more highly
ranked {\sc smooth-shift} transition.\footnote{ If {\sc mitiko} could
represent a topic object in \ex{0}d, there would be another equally
ranked {\sc smooth-shift} interpretation for \ex{0}d.  However,
according to the formulation of {\sc zero topic assignment}, {\sc
mitiko} can not be a zero topic because it was not the $\rm{Cb}$ of
the previous utterance, \ex{0}c.}

These examples clearly show that the {\it wa}-marked {\sc np} is not
always the Cp and support Shibatani's claim that the interpretation
of {\it wa} depends on the discourse context  \cite{Shibatani90}.  The
astute reader will have noticed that in the cases where Hanako is a
zero topic, the {\it wa}-marked Mitiko discourse entity is ranked
according to grammatical function. We conjecture that an
inference of contrast is supported when the grammatical topic is not
the Cp.

The following section discusses the interaction of ZTA with empathy.
Then in section \ref{zta-sum-sec}, we discuss further the
ramifications of our distinction between grammatical and zero topic.

\subsection{Empathy and Zero Topic Assignment}
\label{emp-and-zta-sec}

This section investigates the interaction of {\sc empathy}
and {\sc zero topic assignment} (ZTA) .  The discourse
segment in \ex{1} is a minimal pair with that in
\ex{2}. In \ex{1}d the main verb is {\it setumeisita}
(`explain') without any {\sc empathy} marking, whereas in
\ex{2}d, the same sentence occurs with an auxiliary empathy verb as
{\it setumeisite-kureta}. Remember that {\it kureta} marks the {\sc
obj} or {\sc obj2} as the {\sc empathy locus}.

\eenumsentence{
\item[a.]
\shortex{8}
{Taroo &wa &deeta &o &konpyuutaa &ni &utikondeimasita.}
{Taroo&{\sc top/subj}&data  &{\sc obj} &computer&in&was-storing}
{{\it Taroo was storing the data in a computer.}}

\begin{tabular}{|lll|}
\hline
{\bf Cb:} & {\sc taroo} & \\
{\bf Cf:} & [{\sc taroo}, {\sc data}] & \\ \hline
\end{tabular}

\item[b.]
\shortex{5}
{      0&yatto &hanbun &yari-owarimasita.}
{ {\sc subj}&finally &half &do-finished}
{{\it Finally (Taroo) was half finished.}}

\begin{tabular}{|lll|}
\hline
{\bf Cb:} & {\sc taroo} & \\
{\bf Cf:} & [{\sc taroo}] & {\sc continue} \\ \hline
\end{tabular}

\item[c.]
\shortex{7}
{ Ziroo &ga &0	&hurui &deeta &o &misemasita.}
{ Ziroo&{\sc subj} &{\sc obj2} &old  &data &{\sc  obj} &showed}
{{\it Ziroo showed (Taroo) some old data.}}

\begin{tabular}{|lll|}
\hline
{\bf Cb:} & {\sc taroo} & \\
{\bf Cf1:} & [{\sc taroo}, {\sc ziroo}, {\sc data}] & {\sc zta continue}  \\
           & {\sc top}, {\sc subj}, {\sc obj} & \\  \hline
{\bf Cf2:} & [{\sc ziroo}, {\sc taroo}] & {\sc retain} \\
           & {\sc subj}, {\sc obj2}, {\sc obj}   & \\  \hline
\end{tabular}

\item[d.]
\shortexdt{7}
{      0&  0 &ikutuka &no &kuitigai &o &setumeisimasita.}
{ {\sc subj}&{\sc  obj2} &several &of&differences&{\sc obj} &explained}
{{\it (Ziroo) explained several differences to (Taroo).}}
{{\it (Taroo) explained several differences to (Ziroo).}}

\begin{tabular}{|llll|}
\hline
{\bf Cb1:} & {\sc taroo} & & \\
{\bf Cf1:} & [{\sc taroo}, {\sc ziroo}, {\sc differences}] & {\sc
continue} from Cf1(c) & 12\\
           & {\sc subj}, {\sc obj2}, {\sc obj} & &\\  \hline \hline
{\bf Cb2:} & {\sc ziroo} & &\\
{\bf Cf2:} & [{\sc ziroo}, {\sc taroo}, {\sc differences}] &  {\sc
smooth-shift} from Cf2(c)& 22 \\
           & {\sc subj}, {\sc obj2}, {\sc obj} &  & \\  \hline
\end{tabular}
\label{zta-emp-ga-noemp}
}

The interpretations of \ex{0}d show that it is possible for some
subjects to interpret Taroo as the zero topic in \ex{0}c. This is
possible because Taroo was both the Cp and the Cb for \ex{0}a and
\ex{0}b.  The two Cfs of \ex{0}c reflect multiple possibilities in
attentional state.\footnote{Although both possibilities have the same
semantic interpretation.} The competing defaults consist of the
assumption that ZTA applies, versus the assumption that subjects
are more highly ranked than objects on the Cf. In this case no
preference between the two interpretations can be demonstrated ($Z =
1.79$, not significant).

Example \ex{1} is a minimal pair with \ex{0}.  In \ex{1}d,
the speaker provides more syntactic information by using the
empathy verb {\it kureta} to indicate that the discourse
entity realized as the {\sc object2} is the {\sc empathy}
locus.

\eenumsentence{\item[a.]
\shortex{7}
{Taroo &wa &deeta &o &konpyuutaa &ni &utikondeimasita.}
{Taroo&{\sc top/subj}&data  &{\sc obj} &computer&in&was-storing}
{{\it Taroo was storing the data in a computer.}}

\begin{tabular}{|lll|}
\hline
{\bf Cb:} & {\sc taroo} & \\
{\bf Cf:} & [{\sc taroo}, {\sc data}] & \\ \hline
\end{tabular}

\item[b.]
\shortex{4}
{      0 &yatto &hanbun &yari-owarimasita.}
{ {\sc subj}&finally &half &do-finished}
{{\it Finally (Taroo) was half finished.}}

\begin{tabular}{|lll|}
\hline
{\bf Cb:} & {\sc taroo} & \\
{\bf Cf:} & [{\sc taroo}] & {\sc continue} \\ \hline
\end{tabular}

\item[c.]
\shortex{7}
{ Ziroo &ga &0	&hurui &deeta &o &misemasita.}
{ Ziroo&{\sc subj} &{\sc obj2} &old  &data &{\sc  obj} &showed}
{{\it Ziroo showed (Taroo) some old data.}}

\begin{tabular}{|lll|}
\hline
{\bf Cb:} & {\sc taroo} & \\
{\bf Cf1:} & [{\sc taroo}, {\sc ziroo}, {\sc data}] &  {\sc zta continue}  \\
           & {\sc top}, {\sc subj}, {\sc obj} & \\  \hline
{\bf Cf2:} & [{\sc ziroo}, {\sc taroo}, {\sc data}] & {\sc retain} \\
           & {\sc subj}, {\sc obj2}, {\sc obj}   & \\  \hline
\end{tabular}

\item[d.]
\shortex{7}
{      0&0 &ikutuka &no &kuitigai &o &setumeisite-{\sc kure}-masita.}
{ {\sc subj} &{\sc obj2/emp} &several &of &differences&{\sc obj}
&explained-gave}
{{\it (Ziroo) did (Taroo) a favor of explaining several differences.}}

\begin{tabular}{|llll|}
\hline
{\bf Cb1:} & {\sc taroo} && \\
{\bf Cf1:} & [{\sc taroo}, {\sc ziroo}, {\sc differences}] & {\sc
continue} from Cf1(c) & 33 \\
           & {\sc emp-obj2}, {\sc subj}, {\sc obj} & &\\  \hline \hline
{\bf Cb2:} & {\sc ziroo} && \\
{\bf Cf2:} & [{\sc ziroo}, {\sc taroo}, {\sc differences}] & {\sc
smooth-shift} from Cf2(c) & 1\\
           &  {\sc emp-obj2}, {\sc subj}, {\sc obj} &  & \\  \hline
\end{tabular}
\label{zta-emp-ga}
}

Empathy associates with the previous Cb to yield a {\sc continue}
transition, and the interpretation changes so that the utterance is no
longer ambiguous ($Z = 16.24, p < .001$).  In this case it is possible
to interpret both \ex{0}c and \ex{0}d as {\sc continue}s by assuming
ZTA at \ex{0}c. This example also validates ZTA because empathy
associates with the zero topic \cite{Kuno76a,Kuno87}.  Furthermore,
this minimal pair highlights aspects of the interaction between syntax
and inference.  The fact that the empathy verb in \ex{0}d is the only
difference between \ex{-1} and \ex{0} shows that the preference in
interpretation does not follow from inferences based on information
about who is likely to explain what to whom, depending on who showed
who the data, or whether the data is new or old.

Example \ex{1} contrasts minimally with example \ex{0}
but on another dimension. In this case \ex{1}c is a
{\sc continue} with Taroo realized in subject position,
rather than a {\sc continue} based on ZTA. The
{\it Ziroo explained to Taroo}
interpretation is again clearly preferred here as in
\ex{0}d($Z = 3.638, p < .001$).


\eenumsentence{\item[a.]
\shortex{7}
{Taroo &wa &deeta &o &konpyuutaa &ni &utikondeimasita.}
{Taroo&{\sc top/subj}&data  &{\sc obj} &computer&in&was-storing}
{{\it Taroo was storing the data in a computer.}}

\begin{tabular}{|ll|}
\hline
{\bf Cb:} & [{\sc taroo}]  \\
{\bf Cf:} & [{\sc taroo}, data]  \\ \hline
\end{tabular}

\item[b.]
\shortex{4}
{      0 &yatto &hanbun &yari-owarimasita.}
{ {\sc subj}&finally &half &do-finished}
{{\it Finally (Taroo) was half finished.}}

\begin{tabular}{|lll|}
\hline
{\bf Cb:} & {\sc taroo} & \\
{\bf Cf:} & [{\sc taroo}] & {\sc continue} \\ \hline
\end{tabular}

\item[c.]
\shortex{7}
{ 0&Ziroo &ni&hurui &deeta &o &misemasita.}
{ {\sc subj} &Ziroo&{\sc obj2} &old  &data &{\sc  obj} &showed}
{{\it (Taroo) showed Ziroo some old data.}}

\begin{tabular}{|lll|}
\hline
{\bf Cb:} & {\sc taroo} & \\
{\bf Cf1:} & [{\sc taroo}, {\sc ziroo}, {\sc data}] {\sc continue}& \\
   & {\sc subj}, {\sc obj2}, {\sc obj}   & \\  \hline
\end{tabular}

\item[d.]
\shortex{7}
{0 &  0 &ikutuka &no &kuitigai &o &setumeisite-{\sc kure}-masita.}
{{\sc subj} &{\sc obj2/emp} &several &of &differences&{\sc obj}
&explained-gave}
{{\it (Ziroo) did (Taroo) a favor of explaining several differences.}}

\begin{tabular}{|llll|}
\hline
{\bf Cb1:} & {\sc taroo} & &\\
{\bf Cf1:} & [{\sc taroo}, {\sc ziroo}, {\sc differences}] & {\sc
continue} & 26 \\
           & {\sc emp-obj2}, {\sc subj}, {\sc obj} && \\  \hline
\hline
{\bf Cf2:} & [{\sc ziroo}, {\sc taroo}, {\sc differences}] &  {\sc
retain} & 8 \\
  &  {\sc emp-obj2}, {\sc subj}, {\sc obj} & & \\  \hline
\end{tabular}
\label{zta-emp-cont}
}

In \ex{0} as in \ex{-1},  {\sc empathy}
associates with the previous Cb, ie. Taroo.
We claim that this follows
from the ordering of the Cf and hearers' preferences
for a {\sc continue} interpretation.

Note that the interpretation of the last utterance in \ex{0}d remains
the same as that in \ex{-1}d, although in this case it is Taroo that
shows Ziroo some old data in \ex{0}c; nevertheless Ziroo is the one
who does the explaining. It seems that inference from world knowledge
and domain information alone is unlikely to predict which
interpretations hearers will prefer.  Inferential processes and
discourse structure are mutually constraining
 \cite{JW81,NJ83,Hudson88}.

\subsection{Summary}
\label{zta-sum-sec}

We proposed a discourse rule of {\sc zero topic assignment} and showed
that ZTA is conditioned by the rules and constraints of centering
theory: (1) ZTA only applies to discourse entities that were
previously the Cb; (2) ZTA is constrained to cases where the only
transition available otherwise would be a {\sc retain}.

ZTA arises from the interaction between preferences for {\sc continue}
transitions (Rule 2) and the fact that Cbs are often zeros (Rule 1).
The interaction of these two factors leads to the speculation that
when the Cb is realized by a pronoun in a lower ranked Cf position,
which gives rise to a {\sc retain} transition state, that this type of
transition is inherently ambiguous. Since different factors contribute
to the salience of discourse entities, such as `subjecthood' and
`pronominalization' \cite{GJW86}, conflicting defaults can arise when
these are in conflict with one another. This may be especially true in
Japanese since another factor that should contribute to Cf ranking,
word order, is not present whenever zeros are involved.

These examples highlight the relation between centering and global
coherence in discourse.  A {\sc retain} is proposed as a way for a
speaker to mark a coordinated transition to a new topic; it predicts a
shift \cite{GJW86,BFP87}.  However, the way in which centering {\sc
shift} transitions are related to larger structures in discourse has
not been specified. If a shift functions as a boundary between
segments \cite{Walker93f}, then the hearer's application of ZTA means
that the hearer is assuming that the next utterance will be part of
the same discourse segment. In contrast, a hearer's assumption that
the current centering transition is a {\sc retain} means that the
hearer assumes that the next utterance will begin a new discourse
segment.

The relationship between segmentation and hearer's preferences for ZTA
or {\sc retain} interpretations may be affected by other discourse
factors.  Among these factors, intonation may indicate whether the
current utterance should be taken as initiating a new segment and
predicting a {\sc shift}, or continuing the previous one
\cite{Silverman87,Cahn92,SwertsGeluykens92,WP94}.  Another factor may
be the inferred relationship that holds between adjacent utterances
such as whether it is possible to interpret (d) as Ziroo's reason for
having done (c) \cite{Hobbs85a}. However this is clearly not the only
factor, or even necessarily the dominant one, as we have demonstrated.
Future research must provide additional constraints on when ZTA is
applicable.

\section{Related Research}
\label{rel-res-sec}

Other researchers working on the interpretation of anaphors have
focused on the role of inference from world knowledge
 \cite{Hobbs85a,Hobbs79}.  While it is important to elucidate the
information needed for inference and the type of inferential process
involved in discourse interpretation, it is clear from our examples that
syntactic realization has a strong effect on the interpretive process
and may provide processing constraints on inferential processes. We
have focused on the interaction between syntax and inference.

Our treatment of Japanese discourse phenomena builds on earlier work
by Kuno  \cite{Kuno72,Kuno73,Kuno87,Kuno89}. Our
Cf ranking is consistent with Kuno's Empathy and Topic Hierarchies and
we incorporate a number of Kuno's observations on the function of the
grammatical topic marker {\it wa} and the role of zeros.  We have also
incorporated Kuno's notion of {\sc empathy} by using {\sc empathy} in
the Cf ranking \cite{Kuno76b,Kuno-Kab77}.

In recent work, Kuno proposes an algorithmic account of the
interpretation of zeros. He claims that there are two types of zero
pronouns, {\sc pseudo-zero-pronouns} and {\sc
real-zero-pronouns} \cite{Kuno89}.  {\sc real-zero-pronouns} are
supposed to have a {\it wa}-marked {\sc np} or a presentational {\sc
np} as an antecedent \cite{Yoshimoto88}.  {\sc pseudo-zero-pronouns}
are actually examples of deletion, and must follow the same order and
the same syntactic function as their source {\sc np}s.  They must obey
constraints on deletion such as Kuno's Pecking Order of Deletion
Principle: {\it Delete less important information first and more
important information last}.  According to Kuno, the position just to
the left of the verb is the default focus position in Japanese, unless
the verb itself is the focus. Therefore, since the verb in \ex{1}b is
the information focus, the zeros are assumed to be {\sc
pseudo-zero-pronouns}.

\enumsentence{\item[a.]
\shortex{10}
{Taroo&ga&Hanako&ni&nani&o&sita&no&desu&ka.}
{Taroo&{\sc subj}&Hanako&to&what&{\sc obj}&do&{\sc comp}&{\sc
copula}&{\sc q} }
{{\it What did Taroo do to Hanako?}}

\item[b.]
\shortex{7}
{0 & 0 &kisu&o&sita&no&desu.}
{& & kiss & {\sc obj}&did &{\sc comp}&{\sc copula}}
{(lit.) {\it (Taroo) did a kissing (to Hanako).}}
}

The combination of these two types of zeros can explain examples like
the following:

\eenumsentence{\item[a.]
\shortex{5}
{Taroo &wa &Hanako &ga &sukida.}
{Taroo&{\sc top/subj}&Hanako & & fond-of-is}
{{\it Taroo likes Hanako.}}

\item[b.]
\shortex{5}
{Ziroo &wa &Natuko &ga &sukida.}
{Ziroo&{\sc top/subj}&Natuko & & fond-of-is}
{{\it Ziroo likes Natuko.}}

\item[c.]
\shortexdt{5}
{0 & &Saburoo &mo &sukida.}
{& & Saburoo& also& fond-of-is}
{{\it (Ziroo) also likes Saburoo.}}
{{\it *Saburoo also likes (Natuko).}}
}

Kuno's account treats Ziroo in \ex{0}c as a {\sc real-zero-pronoun}.
In this case we would predict the preferred interpretation based on
our distinction between {\sc continue} and {\sc retain}. However
consider the following example:

\eenumsentence{\item[a.]
\shortex{5}
{Taroo &wa &Hanako &ga &sukida.}
{Taroo&{\sc top/subj}&Hanako & & fond-of-is}
{{\it Taroo likes Hanako.}}

\item[b.]
\shortextt{5}
{Ziroo &wa & &kirai da.}
{Ziroo&{\sc top/subj}& 0 & fond-of-is}
{{\it (Taroo) dislikes Ziroo.}}
{{\it Ziroo dislikes (Hanako).}}
{{\it *Ziroo dislikes (Taroo).}}
}

The {\it Taroo dislikes Ziroo} interpretation would be an example of
ZTA.  However, we would predict that the {\it Ziroo dislikes Hanako}
interpretation would be dispreferred, but this does not seem to be the
case.  Kuno's analysis treats the zero in the second reading of
\ex{0}b as a {\sc pseudo-zero-pronoun} which means that it must be
interpreted as Hanako since Hanako was the object of the previous
utterance.

The interpretation of \ex{0}b that we would predict as possible would
be the {\it Ziroo dislikes Taroo} ({\sc retain}) which native speakers
rarely get. However Kuno's analysis does not block this reading
either; the zero in \ex{0}b could also be a {\sc real-zero-pronoun},
with {\it Taroo\/} as its antecedent.  Kuno says that this
interpretation is dispreferred because of a preference for parallel
interpretation  \cite{GBC78,Sidner79,Kameyama86a,Kuno89}. We have
claimed here and elsewhere \cite{BFP87,WIC90} that the preference for
parallelism is an epiphenomenon of the ordering of the Cf and the
preference for {\sc continue} interpretations.

Our account cannot explain the contrast between \ex{-1} and \ex{0}.
It seems that what is at issue here is the fact that a set of
discourse entities plus an open proposition such as {\it X likes Y} is
what is discourse-old in these examples and not just a discourse
entity \cite{Prince81b,Prince86,Prince92}. Our conclusion is that these
enumerated lists and question-answer discourse segments may need an
account of discourse center that is broader than that needed for
discourse entities realized as {\sc np}s.  Kuno's constraints on
deletion must also be integrated to fully explain when entities or
propositions in the discourse may be unexpressed.

Our analysis also builds on an earlier analysis put forth by Kameyama
 \cite{Kameyama85,Kameyama86b,Kameyama86a}.  Although Kameyama uses the
centering terminology, her account is not based on the constraints and
rules of Centering Theory as developed here and presented in
 \cite{GJW83,GJW86,BFP87}. Kameyama proposed that the interpretation of
zeros in Japanese depends on a default preference hierarchy of
syntactic properties to be shared between the antecedent and the
zero \cite{GBC78}.  Kameyama's account of zero interpretation consists
roughly of a {\sc property-sharing constraint\/}, henceforth PS, and
an {\sc expected center order\/}, henceforth ECO, which may be
paraphrased as follows:

\begin{quote}
{\sc Property-Sharing constraint\/}:
Two zero-pronouns in adjacent utterances, which co-specify the same
Cb-encoding discourse entity, should share one of the following
properties (in descending order of preference): 1) both {\sc ident}
and {\sc subject}, 2) {\sc ident} alone, 3) {\sc subject}-alone, 4)
both {\sc nonident} and {\sc nonsubject}, 5) {\sc nonsubject} alone,
or 6) {\sc nonident} alone.

{\sc Expected Center Order Rule\/}: In a sentence that
contains a center-establishing zero, if it is to have a full
{\sc np} as its antecedent, the default preference order
among its potential antecedent {\sc np}s is: Topic $>$ Ident
$>$ Subject $>$ Object(2) $>$ Others.
\end{quote}

As noted earlier, we use a modified version of Kameyama's {\sc
Expected Center Order} as the ordering of the Cf, but Kameyama's
treatment differs from ours in a number of respects.

First, Kameyama used the property {\sc ident} to describe something
similar to Kuno's notion of {\sc empathy}, and has an added assumption
of a {\sc subject ident} default, i.e.  subjects are consider to be
{\sc empathy} loci by default. This means that her theory also
includes a neutralization device for cases where this default is not
in effect  \cite{Kameyama86a}. In contrast, our theory explains
examples covered by the {\sc subject ident} default by including {\sc
empathy} in the ranking of the Cf list and by the distinction between
{\sc continue} and {\sc retain} as illustrated in example
\ref{emp-cont-ret-ex}.

We have also expanded Kameyama's treatment of {\sc topic}.  We have
elucidated the the interaction of topic with subject and empathy
markers and supported our claim that the topic marker {\it wa}
functions similarly to pronominalization in instantiating the Cb.  In
addition, ZTA and the distinction that we make between grammatical and
zero topic is new to our account.

Furthermore, since Center Instantiation is a side effect of the
application of the Centering algorithm, we treat \ex{1}c and \ex{2}c
with the same mechanism. In Kameyama's analysis, the PS constraint
applies to \ex{1}, while the ECO applies in \ex{2}.

\eenumsentence{
\item[a.]
\shortex{5}{Hanako& wa &repooto&o &kakimasita.}
           {Hanako&{\sc top/subj} &report&{\sc obj} &wrote}
           {{\it Hanako wrote a report.}}

\item[b.]
\shortex{5} {0 & Taroo&ni & aini-ikimasita.}
            {{\sc subj-ident}&Taroo&{\sc obj2}&see-went }
            {{\it She went to see Taroo.}}
\item[c.]
\shortex{5}{Taroo&wa &0&kibisiku&hihansimasita.}
            {Taroo&{\sc top/subj}&{\sc obj} &severely&criticized}
            {{\it Taroo severely criticized her.}}
}

\eenumsentence{
\item[a.]
\shortex{5} {Hanako&wa&Taroo&ni &aini-kimasita.}
            {Hanako&{\sc top/subj}&Taroo&{\sc obj2} &see-came}
            {{\it Hanako came to see Taroo.}}

\item[b.]
\shortex{6} {Taroo&wa & 0 &hon&o &yonde-kure-masita.}
            {Taroo&{\sc top/subj}&{\sc obj2}&book&{\sc obj}&read-gave }
            {{\it Taroo did her a favor of reading a book.}}
}

Note that we annotate \ex{-1}b with Kameyama's {\sc ident} property,
which corresponds to {\sc empathy}.  Kameyama's account predicts that
there are different processes going on in the resolution of zeros
depending on the environments where the zero appears.  PS applies in
\ex{-1}c because the previous utterance has a zero, but doesn't apply
in \ex{0}b. PS would seem to predict that the zero pronoun in \ex{-1}c
should not be interpreted as Hanako, since the zero carries the
properties [{\sc Subj}, {\sc Ident}] in \ex{-1}b and [{\sc NonSubj},
{\sc NonIdent}] in \ex{-1}c.  In other words, {\bf none} of the
required properties of {\sc subj, ident, nonsubj, nonident}, which
`should' be shared according to the PS constraint, are shared.  But in
fact \ex{-1}c is perfectly acceptable under the intended reading of
{\em Taroo severely criticized Hanako} and \ex{0}b is likewise
acceptable under the reading {\em Taroo did Hanako a favor of reading
a book}.

Also, as pointed out in  \cite{Kuno89}, Kameyama's theory makes no
predictions about the interpretation of some of the zeros in examples
such as \ref{cont-ret-ex}, repeated here for convenience as \ex{1}.

\eenumsentence{\item[a.]
\shortex{7}
{Taroo &wa &saisin &no &konpyuutaa &o &kaimasita.}
{ &{\sc  top/subj} &newest &of&computer &{\sc obj} &bought}
{{\it Taroo bought a new model of computer.}}

\begin{tabular}{|lll|}
\hline
{\bf Cb:} & {\sc taroo} & \\
{\bf Cf:} & [{\sc taroo}, {\sc computer}] & \\ \hline
\end{tabular}

\item[b.]
\shortex{7}
{      0 &John &ni &sassoku &sore &o &misemasita.}
{{\sc subj}&John &{\sc obj2}&at once &that &{\sc obj} &showed }
{{\it (Taroo) showed it to John.}}

\begin{tabular}{|lll|}
\hline
{\bf Cb:} & {\sc taroo} &  \\
{\bf Cf:} & [{\sc taroo}, {\sc john}, {\sc computer}] & {\sc
continue} \\ \hline
\end{tabular}

\item[c.]
\shortex{7}
{      0 &0 &atarasiku &sonawatta &kinoo &o &setumeisimasita. }
{{\sc  subj} &{\sc obj2}&newly &equipped &function&{\sc  obj} &explained}
{{\it (Taroo) explained the newly equipped functions to (John).}}

\begin{tabular}{|llll|}
\hline
{\bf Cb:} & {\sc taroo} & &  \\
{\bf Cf1:} & [{\sc taroo}, {\sc john}] &  {\sc continue} & 27  \\
& {\sc subj} {\sc obj}& &  \\  \hline
{\bf Cf2:} & [{\sc john}, {\sc taroo}] &  {\sc retain} & 1 \\
& {\sc subj} {\sc obj} & &  \\  \hline
\end{tabular}
}

The PS Constraint applies only to two zeros in adjacent sentences, and
the ECO applies only when a Cb is to be established.  \ex{0}c is not a
Cb-establishing utterance since the Cb has already been established in
\ex{0}b, so the ECO should not apply. The PS constraint does apply and
predicts that the subject zero must have the subject of \ex{0}b as its
antecedent, but the theory makes no predictions about the possible
interpretations for the zero object.

Many of the examples that are explained in Kameyama's theory by the PS
constraint are handled on our account by the distinction between {\sc
continue} and {\sc retain}.  However, there are a number of cases
where PS makes different predictions than our account. In particular
note that for examples \ref{zta-ex-ga} and \ref{zta-emp-ga},
Kameyama's {\sc subject ident} default makes exactly the opposite
prediction.  \ref{zta-emp-ga} is repeated below as \ex{1} and
annotated with the {\sc subject ident} default feature.

\eenumsentence{\item[a.]
\shortex{7}
{Taroo &wa &deeta &o &konpyuutaa &ni &utikondeimasita.}
{Taroo&{\sc top/subj}&data  &{\sc obj} &computer&in&was-storing}
{{\it Taroo was storing the data in a computer.}}

\item[b.]
\shortex{4}
{      0 &yatto &hanbun &yari-owarimasita.}
{ {\sc subj/ident}&finally &half &do-finished}
{{\it Finally he was half finished.}}

\item[c.]
\shortex{7}
{ Ziroo &ga &0	&hurui &deeta &o &misemasita.}
{ Ziroo&{\sc subj/ident} &{\sc obj2} &old  &data &{\sc  obj} &showed}
{{\it Ziroo showed him some old data.}}

\item[d.]
\shortex{7}
{0&  0 &ikutuka &no &kuitigai &o &setumeisite-{\sc kure}-masita.}
{{\sc subj} &{\sc obj2/ident} &several &of &differences&{\sc obj}
&explained-gave}
{{\it (Ziroo) did (Taroo) a favor of explaining several differences.}}
}

According to PS, the interpretation in which the property {\sc ident}
is shared is preferred to the one with {\sc subject} shared, and
hence, the interpretation {\it Taroo did Ziroo a favor in explaining
several differences} is preferred. However our survey shows that
native speakers prefer the {\it Ziroo did Taroo a favor} reading; this
is explained by our discourse rule of ZTA and by including empathy in
the ranking of the Cf list.

\section{Conclusion and Future Work}
\label{conclusion-sec}

In this paper, we have attempted to elucidate the the interaction of
syntactic realization and discourse salience in Japanese using the
discourse processing framework of {\sc centering}.  In our theory
discourse salience is operationalized by the ranking of the forward
centers for an utterance.  We explored speakers' options for
indicating salience in Japanese discourse, especially the interaction
of discourse markers for {\sc topic} and {\sc empathy}. We then
posited a ranking and used it to explain some facts about the
interpretation of zeros in Japanese.

While there is clearly a correlation between syntax and discourse
function, we show that discourse context plays an important role.  We
proposed a discourse rule of {\sc zero topic assignment} (ZTA) which
distinguishes grammatical and zero topic.  We showed that centering
allows us to formalize constraints on when ZTA can apply.  However
future work must determine additional constraints on when ZTA applies,
and which language families support ZTA.

The preferred interpretation of zeros and the discourse factors which
are responsible for each interpretation are summarized below. Remember
that in each case the zero in the third utterance was established as
the Cb by the previous two utterances:

\begin{center}
\begin{tabular}{|cc|cc|c|c|}
\hline &&&&&\\
\multicolumn{2}{c}{Third Utterance} &
\multicolumn{2}{c}{Fourth Utterance}  &  Discourse  Factor & Example \\
{\sc subject} & {\sc object(2)}& {\sc subject} & {\sc object(2)} &  & \\
\hline  &&&&& \\
zero(i) &    NP(j)  &          zero(i) &   zero(j) &         Continue/Retain &
 \ref{cont-ret-ex}\\
\hline  &&&&& \\
zero(i)   &  NP(j) &      zero(j) &   zero(i),empathy  &   empathy,
Continue/Retain & \ref{zta-emp-cont} \\
\hline  &&&&&\\
NP(ga)(i) & zero(j) & zero(j) & zero(i) & ZTA & \ref{zta-ex-ga},
\ref{zta-emp-ga-noemp}\\
\hline  &&&&& \\
NP(wa)(i)  &  zero(j)  &        zero(i)  &  zero(j)       &      WA-effect&
\ref{zta-ex-wa} \\
           &           &        zero(j)  &  zero(i)         &    ZTA \\
\hline  &&&&& \\
NP(ga)(i)  &  zero(j)    &      zero(i) &   zero(j),empathy  &   ZTA and
empathy & \ref{zta-emp-ga}\\
\hline
\end{tabular}
\end{center}

This analysis suggests that centering may be a universal of
context-dependent processing of language, although so far this theory
has only been applied to English, German, Japanese, Italian and
Turkish
\cite{BFP87,Walker89b,WIC90,Dieugenio90,Cote92,Rambow93,Nakatani93a,Hoffman95,Turan95}.
We proposed that the centering component of a theory of discourse
interpretation can be constructed in a language independent fashion,
up to the declaration of a language-specific value for one parameter
of the theory, i.e., Cf ranking (as in section \ref{cent-sec}).  This
parameter is language-dependent because different languages offer
different means of expressing discourse function. We conjecture that
ZTA may apply in any free-word order language with zeros.

Future work must examine the interaction between centering and
discourse segmentation in both monologue and dialogue
\cite{WS88,WW90,Walker93f}, and the role of deictics, lexical semantics,
one anaphora, and propositional discourse entities in centering
\cite{Webber78,Sidner79,Walker92a,Walker93c,Cote95}.  It is also
important to examine the interaction of zeros with overt pronouns and
with deictics, and the interaction of pronominalization with accenting
\cite{Terken95}.  In addition, the semantic theory underlying
centering must be further developed \cite{Roberts95}.  Finally,
centering transitions are currently defined by an equality relation
between discourse entities, but poset relations and functional
dependencies often link entities in discourse
\cite{Prince78a,Prince81b,Ward85,GJW86}.  The predictions made here
should also be tested on a large corpus of naturally occurring
Japanese discourse \cite{HL95}.

\section{Acknowledgements}

We'd like to thank Aravind Joshi and and Ellen Prince for their
insight, useful discussions and support, and for making it possible to
have the Workshop on Centering Theory in Naturally-Occurring Discourse
at the Institute for Cognitive Science at the University of
Pennsylvania.  In addition, discussions with Dave Bernstein, Susan
Brennan, Hitoshi Isahara, Megumi Kameyama, Susumu Kuno, Christine
Nakatani, Hiday Nakashima, Carl Pollard, Owen Rambow, Peter Sells,
Mike Tanenhaus, Bonnie Webber and Steve Whittaker contributed to the
development of this work.  We'd also like to thank the anonymous
reviewers who provided many helpful suggestions. NSF's Summer Science
and Engineering Institute in Japan made it possible to present this
work and receive useful feedback at ICOT, JEIDA working group on
Machine Translation, NTT's Basic Research Labs, and ATR's Interpreting
Telephony Lab.  We would also like to thank the readers of
comp.research.japanese and sci.lang.japan who participated in our
survey.

This research was partially funded by NSF Science and Engineering
Award for the Summer Institute in Japan, by ARO grant
DAAL03-89-C-0031, DARPA grant N00014-90-J-1863, NSF grant IRI90-16592,
and Ben Franklin grant 91S.3078C-1 at the University of Pennsylvania,
and by Hewlett Packard Laboratories.

\section{Appendix: Instructions to Survey Participants}

\subsection{Instructions for Survey 1 and 2}

What interpretation do you get for the THIRD sentence
of each set where there are two unexpressed arguments?
0(i) in the second sentence indicates that the
unexpressed argument in the sentence should be
interpreted as referring to the {\sc np} of the first
sentence marked with (i).  Please rank your preference:
it's ok to have more than one equally preferred
interpretation.

\subsection{Instructions for Survey 3}

Dear Participants.  Thank you for serving as subjects
for us for this informal experiment.  You can help us
most by following the directions here.  Please read
each sample discourse in turn and make your
interpretation as rapidly as possible. Do not scroll
back and forth in the file.  Please indicate which
interpretation, (a) or (b) you get by marking your
choice with a 1.  It is very important that you choose
*one* interpretation only, and the one you choose
should be the first one that you think of as you are
reading the sample discourse.  Send us back this file
with your choices marked.


\end{document}